\documentclass{article}
\usepackage[utf8]{inputenc}
\usepackage[english]{babel}
\usepackage[top=2cm,bottom=2cm,left=3cm,right=3cm]{geometry}

\usepackage{amsmath,amsfonts,amssymb,amsthm}
\usepackage{mathtools}
\usepackage[svgnames]{xcolor}
\usepackage{graphicx}
\usepackage{braket}
\usepackage{array}
\usepackage{numprint}
\usepackage{tikz}
\usepackage{adjustbox}
\usepackage{subcaption}
\usepackage{aliascnt}
\usepackage{pgfplots}
\usepackage{bbm}
\usepackage{numprint}
\usepackage[color=LightSkyBlue]{todonotes}

\usepackage{algorithm}
\usepackage{algpseudocode}
\usepackage{csquotes}

\usepackage{enumerate}
\usepackage[style=apa, bibencoding=auto,backend=biber,natbib]{biblatex}
\usepackage[hidelinks,colorlinks=true,linkcolor=blue,citecolor=blue]{hyperref}

\pgfplotsset{
standard/.style={
    axis x line=middle,
    axis y line=middle,
    enlarge x limits=0.15,
    enlarge y limits=0.15,
    every axis x label/.style={at={(current axis.right of origin)},anchor=north west},
    every axis y label/.style={at={(current axis.above origin)},anchor=north east},
    every axis plot post/.style={mark options={fill=white}}
    }
}
\pgfplotsset{compat=1.17}

\DeclareMathOperator*{\argmax}{arg\,max}

\DeclareMathOperator*{\pr}{Pr}
\DeclareMathOperator*{\normalgamma}{NormalGam}
\DeclareMathOperator*{\invgamma}{InvGam}
\DeclareMathOperator*{\Log}{log}
\DeclareMathOperator*{\Gam}{Gam}
\DeclareMathOperator*{\Regret}{Regret}

\newcommand{\prob}[1]{\pr\left(#1\right)}
\newcommand{\condpar}[2]{\left(#1 \;\middle|\; #2\right)}
\newcommand{\condsquare}[2]{\left[#1 \;\middle|\; #2\right]}
\newcommand{\condprob}[2]{\pr\left(#1 \;\middle|\; #2\right)}
\newcommand{\expect}[1]{\mathbb{E}\left[#1\right]}
\newcommand{\condexpect}[2]{\mathbb{E}\condsquare{#1}{#2}}

\newcommand{\pre}{\tau}
\newcommand{\Exp}[1]{\exp\left(#1\right)}

\newcommand{\transpose}{\mathsf{T}}
\newcommand{\precision}{\Lambda}
\newcommand{\samplemean}{\hat{\mu}}

\newcommand{\probc}{\zeta}

\newcommand{\rdim}{d}

\newtheorem{thm}{Theorem}[section]
\newaliascnt{conj}{thm}

\newaliascnt{prop}{thm}

\newaliascnt{coro}{thm}

\newaliascnt{lem}{thm}
\newtheorem{lemma}[lem]{Lemma}

\newtheorem*{mainlemma}{Main Lemma}

\theoremstyle{definition}
\newtheorem{definition}{Definition}[section]

\addto\extrasenglish{%
}

\title{Thompson Sampling for Linear Bandit Problems with Normal-Gamma Priors}
\author{Björn Lindenberg, Karl-Olof Lindahl}
\begin{document}
\maketitle
\begin{abstract}
We consider Thompson sampling for linear bandit problems with finitely many independent arms, where rewards are sampled from normal distributions that are linearly dependent on unknown parameter vectors and with unknown variance. Specifically, with a Bayesian formulation we consider multivariate normal-gamma priors to represent environment uncertainty for all involved parameters. We show that our chosen sampling prior is a conjugate prior to the reward model and derive a Bayesian regret bound for Thompson sampling under the condition that the 5/2-moment of the variance distribution exist.  
\end{abstract}
\section{Introduction}
We consider a learner or an agent that is given a set of actions. Upon executing an action the agent observes a reward drawn randomly from some associated probability distribution for that action. Given uncertainty about the nature of the reward system, the agent must sequentially select actions, observe rewards and learn how to maximize the total payoff in expectation. Thus, the agent must strike a balance between \emph{exploring} less understood actions and \emph{exploiting} posterior knowledge for greater gain. In this regard, a common performance measure is for agent algorithms to be judged by the \emph{Bayesian regret} or \emph{Bayesian risk}. For a finite time horizon the measure is defined to be the expectation of accumulated differences in gain between optimal actions and chosen actions, when viewed to incorporate both uncertainty in environments, randomness in outcomes and randomness in agent policies. 

Numerous studies such as by \textcite{auer2002using} are focused on algorithms that explicitly utilize the principle of \emph{optimism in the face of uncertainty}. The core idea is to provide an estimation of the upper confidence bound of the expected reward for each arm, and then always choose the action which maximizes the bound. However another approach, which we study in this paper, is \emph{Thompson sampling} (TS). The sampling procedure draws estimations of expected rewards from the posterior distribution over parameters conditioned on observed rewards and a modeled prior distribution. It then selects the best action according to the sampled set of estimations. Hence TS maintains an uncertainty of the current environment with growing confidence of optimality as more rewards are observed and digested by the prior. Moreover by an argument of confidence bounds, \textcite{russo2014learning} showed that TS preserves the core principle without an explicitly designed optimism.

Even if the sampling procedure was first proposed in \textcite{thompson1933likelihood}, it took until recent times for it to attain strong provable guarantees, in particular its asymptotical convergence to optimality (see for example \textcite{may2012optimistic}). It should also be noted that convergence, in its strictest sense, requires that every action is sampled infinitely often. This is not always the case when a TS algorithm uses a prior belief that is \emph{misspecified} against the true underlying \emph{Bayesian prior} of environment uncertainty. Moreover, even in the case of convergence, misspecification may also lead to an excess of explorative actions or to the agent being biased towards non-optimal actions. In both cases the learning performance suffers (see for example \cite{russo2018tutorial}).

In this paper we study a correctly specified TS algorithm for a generalized class of problems called \emph{contextual bandits}, where conventional TS algorithms are almost surely misspecified. Specifically, we look at \emph{stochastic linear bandits} with finitely many independent arms introduced by \textcite{abe1999associative}, where rewards are assumed to be normally distributed and linearly dependent on unknown parameters. Conventional models here assume a Gaussian Bayesian prior for the parameters, but with fixed noise variance. Given a \emph{normal-gamma} Bayesian prior, we further extend this model by considering bandits with \emph{uncertainty on both parameters and noise variance}, i.e., we consider \emph{normal-gamma linear bandits}. We show that our chosen TS prior is a conjugate prior to the reward model, which leads to well-defined updates for the posterior sampling distribution. 

Our main contribution is a Bayesian regret bound for TS on normal-gamma bandits. For a finite action set of size $K$ and arbitrary priors over bounded reward distributions, \textcite{russo2014learning} showed that TS for a finite time horizon $T$ attains a regret bound of order $\sqrt{K T \Log T}$. With unbounded rewards and sufficiently well-behaved variance distributions, we attain a similar but sharper regret bound of order $\sqrt{K T W_0(T/K^\xi)}$, $\xi \in (0,1)$, where $W_0$ is Lambert's principal $W$-function satisfying $W_0(T) = o(\Log T)$.

\subsection{Related Work}
\textcite{scott2010modern} and \textcite{chapelle2011empirical} provided strong empirical evidence for the ease-of-use and competitiveness of Thompson Sampling for contextual bandits. However, general finite-time horizon bounds for the Bayesian regret remained limited and the development of further bounds was raised as an open question. This was remedied by \textcite{russo2014learning}, who provided the general TS Bayesian regret bound of order $\sqrt{K T \Log T}$ (see also \textcite{slivkins2019introduction}). The result assumes arbitrary Bayesian priors on bounded rewards. In contrast, our work with a fixed Bayesian prior considers unbounded rewards and unbounded parameter sets.

Further progress can be made given rewards that are contained in $[0,1]$ and bandits defined by mean reward vectors. \textcite{bubeck2013prior} remove the extraneous logarithmic factor and attains a TS Bayesian regret bound of order $\sqrt{K T}$ for arbitrary bounded priors. For the same class of bandits but with a fixed Gaussian TS prior, \textcite{agrawal2017near} provides near optimal bounds of order $\sqrt{K T \Log K}$ regardless of reward distributions. In contrast, our work assumes linear bandits under normal-gamma Bayesian priors with unbounded rewards. 

Only recently have initial steps been taken towards the study of bandits that incorporate risk. \textcite{audibert2009exploration} study algorithms that use variance estimates for upper confidence bounds and show that such algorithms may have an advantage over non-estimating alternatives. \textcite{vakili2015mean} study risk-aware learning policies with strong theoretical results for the non-Bayesian regret. \textcite{zhu2020thompson} consider for the first time Thompson sampling in the context of mean-variance optimization for bandits defined by Gaussian reward distributions. In their work they model normal-gamma TS priors for univariate mean rewards and risk, and provide non-Bayesian regret bounds expressed in terms of fixed environment parameters. In contrast, our work considers linear bandits with environment uncertainty under a Bayesian prior. Moreover, we provide an additional analysis on the corresponding Bayesian regret. 

\subsection{Organization}
The rest of this paper is organized as follows. In \autoref{sec:problem} we make a formal introduction to TS and the normal-gamma bandit. In \autoref{sec:mainresult} we present our main contribution with a Bayesian regret bound and the main lemma used to prove the result. In \autoref{sec:tsng} we then explicitly derive formulas for TS posterior updates given a normal-gamma prior belief and present the corresponding TS algorithm. We then conclude the section by presenting an empirical study of the effects of misspecification with a comparison of normal-gamma TS versus a Gaussian TS method. To aid with analysis, we look at implied distributions for TS prior parameters in \autoref{sec:distbounds}, and present asymptotical bounds for random variables related to model parameters under the Bayesian prior. In \autoref{sec:regretanalysis} we prove our main result providing a Bayesian regret bound under the normal-gamma environment uncertainty. We conclude with some final remarks and future work in \autoref{sec:discussion}.

\section{Problem Formulation}
\label{sec:problem}
\subsection{Stochastic Linear Bandits}
\label{sec:slb}
Using the framework of \textcite{lattimore2020bandit}, we consider \emph{contextual linear bandits}. For each time step $t \in \mathbb{N}$, a learner has access to a finite decision set $\mathcal{A}_t \subset \mathbb{R}^d$ from which a $d$-dimensional \emph{context} vector $A_t = a \in \mathcal{A}_t$ is chosen. The learner then observes a random reward
\begin{equation}
\label{eq:stdlinearreward}
    X_t = a^\transpose \theta + \eta(t),
\end{equation}
where $\theta \in \mathbb{R}^d$ is some unknown vector of model parameters and $\eta(t)$ is 1-sub-Gaussian noise. 

In the case when for all $t$, $\mathcal{A}_t = \mathcal{A}$ and $\eta(t) = \eta$, we have a \emph{static linear bandit}, which is a restriction placed on the bandits in this paper. Expanding these notions further, the model parameters $\theta$ and the noise variable $\eta$ may not be the same for each action  (\textcite{slivkins2019introduction}). Hence if we let $K \coloneqq \left|\mathcal{A}\right|$ and put $[K] \coloneqq \set{1, 2, \dots, K}$. Then we may identify each action by $a_k$, $k \in [K]$, and expand the reward definition by
\begin{equation}
\label{eq:lineareward}
    X_t^{(k)} = a_k^\transpose \theta_k + \eta_k,
\end{equation}
where the noise variables $\left(\eta_k\right)_{k = 1}^K$ are assumed to be independent. Thus, model parameters here contains the $K \times d$-dimensional parameter $\text{Vec}\left(\left[ \theta_1, \theta_2, \dots, \theta_K\right]\right)$. We note that the formulation of \eqref{eq:lineareward} can be seen as a restriction of \eqref{eq:stdlinearreward} since we can always embed each context $a_k$ in a larger space with padded zeros. In many cases $\eta$ is modeled by a fixed noise distribution, e.g., $\mathcal{N}(0, \sigma^2)$. However, in the following treatment we will assume that the distribution of $\eta_k$ is parameterized but possibly unknown, and in that case we let the corresponding parameters be included in the total vector $\theta$ of model parameters.

With a Bayesian formulation, we may put an uncertainty on $\theta$ over some subset $\Theta$ of real vectors. The distribution of $\theta$ is called the \emph{Bayesian prior} and encapsulates the agent's inability to beforehand know the true nature of the system, i.e., in this formulation each realized $\theta$ corresponds to a possible problem instance $\mathcal{E}(\theta)$. Let $n_k(t)$ be the random variable which after $t$ steps counts the number of times action $a_k$ was selected. For each action we let $\mu_k \coloneqq \condexpect{X_1^{(k)}}{\theta}$ denote the expected mean reward. An optimal action $A^*$ conditioned on $\theta$ is then any action indexed in $\argmax_{k \in [K]} \mu_k$ with the corresponding optimal mean reward $\mu^*$, and any deviation from an optimal action yields an \emph{immediate regret} $\Delta_k \coloneqq \mu^* - \mu_k$.

A common measure of an agent's performance is then the random variable
\begin{equation}
\label{eq:regret}
    \Regret(T, \theta) \coloneqq \condexpect{\sum_{t = 1}^T \Delta_{A_t}}{\theta} = \condexpect{\sum_{k \in [K]} n_k(T) \Delta_k}{\theta},
\end{equation}
which benchmarks the expected cumulative sum of immediate regrets for a finite time horizon $T$.
Since there is uncertainty in $\theta$ we therefore have the \emph{Bayesian regret} or the \emph{Bayesian risk}
\begin{equation}
\label{eq:bayesregret}
    \Regret(T) = \expect{\Regret(T, \theta)} = \expect{\sum_{k \in [K]} n_k(T) \Delta_k},
\end{equation}
which is taken as the expected regret over all problem instances. Hence minimization of \eqref{eq:bayesregret}, in the context of Bayesian bandits, is a common goal for algorithms. 

\subsection{Thompson Sampling}
First proposed by \textcite{thompson1933likelihood}, Thompson sampling is an algorithm that involves a learner that initially chooses a prior $\mathbb{P}_0$ over possible bandit environments. In each round the algorithm samples an environment from the maintained posterior distribution $\mathbb{P}_{t-1}$ given an observed history $A_1, X_1, \dots, A_{t-1}, X_{t-1}$ of actions and outcomes. It then acts according to the optimal action $A_t$ for the environment, observes an outcome $X_t$ and updates the posterior by Bayes' rule. Formally for environments indexed by a parameter set $\Theta$ we have the following algorithm:
\begin{algorithm}[H]
\caption{Thompson Sampling}\label{alg:ts}
\begin{algorithmic}
\Require Action set $\mathcal{A}$, distribution $\mathbb{P}_0$ over $\Theta$
\For{$t = 1, 2, \dots T$}
\State Sample $\Tilde{\theta}\sim \mathbb{P}_{t-1}$ 
\State Choose $A_t \in \argmax_{a \in \mathcal{A}} \mu_a(\Tilde{\theta})$
\State Observe $X_t$
\State Update $\mathbb{P}_t \gets \mathbb{P}\left(\cdot \mid A_1, X_1, \dots, A_t, X_t\right)$
\EndFor
\end{algorithmic}
\end{algorithm}
\noindent Thompson sampling is known to be asymptotically close to optimal in a variety of settings (\textcite{lattimore2020bandit}). The learner may explore efficiently by the randomization procedure, given that the prior displays a large enough spread of initial concentration. As data is gathered, the posterior then concentrates around the true parameter-point and the exploration rate decreases. However, the choice of prior can have a significant effect on performance (\textcite{ lattimore2020bandit}). If the prior is sufficiently \emph{misspecified} there is a possibility that the algorithm catastrophically underestimates the optimal arm and never plays it. Whereas distributions that are chosen to be \emph{coherent}, i.e., of the same form as the true distribution of environments, removes this issue along with superior learning performance versus misspecification (\textcite{russo2018tutorial}).

\subsection{Normal-Gamma Linear Bandits}
\label{sec:nglb}
Consider a linear bandit problem where the indexing set of actions is $[K] \coloneqq \set{1, 2, \dots, K}$. We recall that the agent for each time step $t$ has access to a \emph{decision set} $\mathcal{A} = \set{a_1, a_2, \dots, a_K} \subset \mathbb{R}^d$ for some dimension $d$. If the reward for each arm is given by \eqref{eq:lineareward}, with the assumption of $\eta_k \sim \mathcal{N}(0, \sigma^2)$ for all arms. Then we have a \emph{Gaussian linear bandit} where the reward for action $a_k$ is distributed as
\begin{equation}
\label{eq:gausslinear}
    X_t^{(k)} \sim \mathcal{N}(a_k^\transpose \theta, \sigma^2).
\end{equation}
Models found in literature, often take $\theta_k = \theta$ to be the same set of parameters for all $k$ and uncertainty is modeled by $\theta \sim \mathcal{N}_d\left(\theta_0, \Sigma_0\right)$ for some mean $\theta_0$ and covariance $\Sigma_0$ (see \textcite{agrawal2013thompson, russo2018learning, hao2021information}).

As stated in \autoref{sec:slb}, we may assume distinct action parameters, with a different linear function and noise variance for each arm. We may further assume that the Gaussian noise variance is unknown, i.e., part of the unknown parameter vector $\theta$. Thus, we now proceed to present the \emph{normal-gamma linear bandit} studied in this paper, where $\sigma^2$ in \eqref{eq:gausslinear} is assumed to be unknown, distinct for each arm and in the form of a precision value $\tau$.

Explicitly, we let the model be defined by parameters $(\theta_k, \tau_k) \in \mathbb{R}^d \times \mathbb{R^+}, k \in [K]$, where the reward $X^{(k)}_t$ for choosing $a_k \in \mathcal{A}$ is distributed as 
\begin{equation}
\label{eq:likelihood}
    X_t^{(k)} \sim \mathcal{N}\left(a_k^\transpose \theta_k , \left(\tau_k \right)^{-1}\right).
\end{equation}
We recall that $a_k$ can be seen as a \emph{context} of $d$ features for the $k$th action, which when deriving the reward distribution is subjected to the linear map $\langle \cdot, \theta_k \rangle \colon \mathbb{R}^d \to \mathbb{R}$ with added Gaussian noise involving some unknown precision $\tau_k$. Further, let the Bayesian prior of $(\theta, \tau)$ for any arm be defined as follows. Given some real vector $\theta^* \in \mathbb{R}^d$, a positive definite $d \times d$ matrix $\precision^*$, and scalars $\alpha^*, \beta^* \in \mathbb{R}^+$, we assume that
\begin{equation}
\label{eq:prior}
\begin{aligned}
        \tau &\sim \Gam(\alpha^*, \beta^*), \\
    \theta \mid \tau &\sim \mathcal{N}_d \left(\theta^*,  \left(\tau \precision^*\right)^{-1}\right).
\end{aligned}
\end{equation}
Thus, a realized environment can be seen as a sampled sequence of parameters $\left( (\theta_k, \tau_k) \right)_{k=1}^K$, which are independent and drawn from  normal-gamma distributions $\left(\normalgamma(\theta_k^*, \precision_k^*, \alpha_k^*, \beta_k^*)\right)_{k = 1}^K$. 
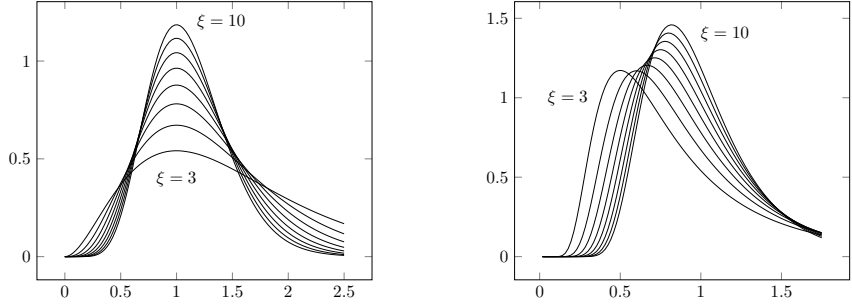
\begin{figure}[t]
    \centering
    \begin{subfigure}[b]{0.45\textwidth}
        \hfill
        \begin{tikzpicture}[scale=0.65]
        \begin{axis}
        \foreach \k in {3, ..., 10} {
            \addplot[domain=0:2.5,samples=100]{(\k-1)^\k*1/factorial(\k-1)*x^(\k-1)*exp(-(\k-1)*x)};
        }
        \node at (axis cs:1,0.4) {$\xi=3$};
        \node at (axis cs:1.4,1.2) {$\xi=10$};
        \end{axis}
        \end{tikzpicture}
    \end{subfigure}
    \hfill
    \begin{subfigure}[b]{0.45\textwidth}
        \begin{tikzpicture}[scale=0.65]
        \begin{axis}
        \foreach \k in {3, ..., 10} {
            \addplot[domain=0:1.75,samples=100]{(\k-1)^\k*1/factorial(\k-1)/x^(\k+1)*exp(-(\k-1)/x)};
        }
        \node at (axis cs:0.175,1.0) {$\xi=3$};
        \node at (axis cs:1.15,1.4) {$\xi=10$};
        \end{axis}
        \end{tikzpicture}
    \end{subfigure}
    \hfill
    \caption{Densities of $\Gam(\xi, \xi-1)$ (left) and $\invgamma(\xi, \xi-1)$ (right) for $\xi = 3, 4, \dots, 10$, where the expectation of the inverse in each case is 1.}
    \label{fig:gamma}
\end{figure}
In the case when $\theta_k^* = 0$, $\alpha_k^* = \alpha^*$ and $\beta_k^* = \beta^*$ for all $k \in [K]$, we obtain iid precision sequences $\tau_1, \tau_2, \dots, \tau_K$ drawn from $\Gam(\alpha^*, \beta^*)$. In addition, the reward centers $\mu_k \coloneqq a_k^\transpose \theta_k$ are then conditionally distributed as
\begin{equation*}
    \mu_k \mid \tau_k \sim \mathcal{N}\left(0, 1/(\lambda_k^* \tau_k)^{-1}\right),
\end{equation*}
where $\lambda_k^* \coloneqq \left(a_k^\transpose \left(\precision_k^*\right)^{-1} a_k\right)^{-1}$ is fixed and defined by the context $a_k$ and the hyperparameter $\precision_k^*$. It follows that the immediate regret $\Delta_k$ conditioned on drawn precisions can be seen as the random variable $\max_i \mu_i - \mu_k$ comprised of conditional Gaussian variables. So given these remarks we will restrict our analysis to arm-independent Bayesian priors
\begin{equation*}
    \normalgamma\left(0, \precision_k^*, \alpha^*, \beta^* \right), \ k \in [K],
\end{equation*}
where the precision distribution $\Gam(\alpha^*, \beta^*)$ is the same for each $k \in [K]$ and where $\theta_k^* = 0$ such that $\mu_k$ is centered around 0. Moreover, for a more well-behaved inverse distribution, we will assume that the shape parameter satisfies $\alpha^* > 5/2$ where both the mean and variance of the inverse exist (see \autoref{fig:gamma}).
\section{Main Result}
\label{sec:mainresult}
We now present the main result of this paper, which is a TS-regret bound for the normal-gamma linear bandits presented in \autoref{sec:nglb} under the assumption of finitely many independent arms with $\normalgamma\left(0, \precision_k^*, \alpha^*, \beta^* \right)$ Bayesian priors for each arm.
\begin{thm}[Regret Bound]
\label{thm:main}
Let contexts $a_1, a_2, \dots, a_K \in \mathbb{R}^d$ be normalized. If model parameters $\theta = \left(\theta_k, \tau_k \right)_{k = 1}^K$ consists of independently drawn $(\theta_k, \tau_k)$ from distributions
\begin{equation*}
  \normalgamma\left(0, \precision_k^*, \alpha^*, \beta^* \right),
\end{equation*}
$\alpha^* > 5/2$, for each arm $k \in [K]$. Then for sufficiently a large time horizon $T$ the Bayesian regret for TS is
\begin{equation*}
    \mathcal{O}\left(\sqrt{K T W_0\left(\displaystyle \frac{T}{K^{1 - 2\epsilon}}\right)} \right),
\end{equation*}
where $1/\alpha^* < \epsilon < 2/5$ and $W_0(x)$ is Lambert's principal W-function satisfying $W_0(x) < \Log x$ for $x > e$. 
\end{thm}
It should be noted that the Bayesian regret in \autoref{thm:main} holds for reward and parameter distributions with unbounded support. Moreover, compared to the well-known general TS-bound $\sqrt{K T \Log T}$ from (\textcite{russo2014learning, slivkins2019introduction}), dependence on $\Log T$ is replaced by the slower growing Lambert function $W_0(x)$.

\subsection{Statement of the Main Lemma}
The main result is a consequence of the following definition and lemma conditional on a sampled environment with parameters $\theta = \left(\theta_k, \tau_k\right)_{k=1}^K$.
\begin{definition}
\label{def:cd}
For a sampled set of model precisions $\left(\tau_k \right)_{k = 1}^K$, put $\tau_0 \coloneqq \min_k \tau_k$. We then define $\tau_0$-dependent quantities $C, D$ as follows
\begin{align*}
    C(\tau_0) &\coloneqq \frac{2}{\tau_0} \sqrt{\frac{ \tau_0}{2} + \Log 2} + \frac{1}{\tau_0} \left(2 \Log 2 + 1 \right) + 3, \\
    D(\tau_0) &\coloneqq \frac{8}{\tau_0}\left(1 + \sqrt{\frac{\tau_0 \left(2 C(\tau_0) + \frac{1}{4} \right)}{2}} \right)^2.
\end{align*}
\end{definition}

\begin{mainlemma}[Action Regret]
\label{thm:mainlemma}
Given model parameters $\theta = \left(\theta_k, \tau_k\right)_{k=1}^K$, put $\tau_0 \coloneqq \min_k \tau_k$ and let $D = D(\tau_0)$. Let $n_k(T)$ be defined as the random variable which after $T$ steps counts the number of times $a_k$ was selected. Then for all $T \geq 2$,
\begin{equation*}
    \condexpect{n_k(T)}{\theta} \Delta_k < 
            \frac{7 D}{2} \left(\Delta_k + \frac{1}{\Delta_k}\right) + 2D\left(\Log D \Delta_k + \frac{1}{\Delta_k} \Log \left(\frac{D}{\Delta_k^2} \right)\right) + 9 \Delta_k.
\end{equation*}
\end{mainlemma}
\noindent The \hyperref[thm:mainlemma]{Main Lemma} shows a bound for the conditional regret of each arm in terms of the parameter dependent value $D(\tau_0)$ and the immediate regret $\Delta_k$. An initial starting point for its derivation is by considering expectations of probabilities for high and low probability threshold events. The value of $D(\tau_0)$ is then an artifact from a procedure that bounds these expectations in terms of a common quantity. The bound can ultimately be viewed as a sum of two parts; one part containing potentially unbounded non-reciprocal values $\Delta_k$ and one part containing potentially unbounded reciprocal values $1/\Delta_k \Log(1/\Delta_k^2)$. \autoref{thm:main} then follows from analyzing the expectations of each part under the Bayesian prior.
% \todo{Att överväga: ska man lägga till kommentar om $\tau_0$ och high probability events här?}
% \todo[inline, color=orange]{Fixed. Finns ingen enkel referens längre för sättet är halv-unikt. En del är inspirerat av AgrawalGoyal, som reffas och återbevis senare. Den andra delen med ett uttryck av low probability threshold events är unik. "Standard approach" som var skrivet där nere var bara en gammal kvarleva. Tror inte $D$ eller $\tau_0$ betyder något mer än att det var enklaste sättet för att hitta en gemensam kvantitet $\rho_k$ så man kunde knyta ihop gränser för summor innehållandes $p_k$ och $q_k$. Sättet var långt ifrån optimalt för att hitta gränser, mest att det blev analytiskt enkelt. Fast future work och körningar visar ändå att tillräckligt fångas för att det asymptotiskt ser rätt ut trots det $\sim \log^2 T$.}

\section{Thompson Sampling with the Normal-Gamma Prior}
\label{sec:tsng}
We now proceed to describe the Thompson sampling updates for the normal-gamma linear bandit. Focusing on a single arm, we let $(\theta, \tau) \coloneqq (\theta_k, \tau_k)$, put $a \coloneqq a_k$ and set prior normal-gamma parameters $(u, \precision, \alpha, \beta)$ for our uncertainty about $(\theta, \tau)$. 

By \eqref{eq:likelihood} we have the likelihood
\begin{equation}
\label{eq:likedensity}
    L\condpar{x}{\theta, \tau} = \frac{\tau^{\frac{1}{2}}}{\sqrt{2 \pi}} \exp\left(-\frac{\tau}{2} \left(x - a^\transpose \theta \right)^2\right).
\end{equation}
Moreover, by $\eqref{eq:prior}$ we have the $(\theta, \tau)$-density
\begin{equation}
\label{eq:priordensity}
\begin{aligned}
        &f\condpar{\theta, \tau}{u, \precision, \alpha, \beta}\coloneqq f\condpar{\theta}{\tau, u, \precision}f\condpar{\tau}{\alpha, \beta} \\
&=\frac{\sqrt{\left|\precision\right| }\beta^\alpha \tau^{\alpha - 1 + \frac{d}{2}}}{\left(2 \pi\right)^{\frac{d}{2}} \Gam(\alpha)}  \exp\left(-\frac{\tau}{2}\left(\theta - u \right)^\transpose \precision \left(\theta - u\right) - \beta \tau \right).
\end{aligned}
\end{equation}
The posterior now follows from the fact that it is proportional to $L\condpar{x}{\theta, \tau}f\condpar{\theta, \tau}{u, \precision, \alpha, \beta}$ and the fact that $f\condpar{\theta, \tau}{u, \precision, \alpha, \beta}$ is a conjugate prior to $L\condpar{x}{\theta, \tau}$. Explicitly we have the following result.
\begin{lemma}
\label{thm:updates}
The parameters of the posterior $\normalgamma(u', \precision', \alpha', \beta')$ given an observation $x$ and prior $\normalgamma(u, \precision, \alpha, \beta)$ can be stated as
    \begin{align*}
        u' &= \left(\precision + a a^\transpose \right)^{-1} \left(x a + \precision u \right), \\
        \precision' &= \precision + a a^\transpose, \\
        \alpha' &= \alpha + \frac{1}{2}, \\
        \beta' &= \beta + \frac{1}{2} \left( x^2 + u^\transpose \precision u- u'^\transpose \precision' u'\right).
    \end{align*}
\end{lemma}

\begin{proof}
With a conjugate prior ansatz and the added information of $x$ we assume that
\begin{equation*}
    f\condpar{\theta, \tau}{x, u', \precision', \alpha', \beta'} \propto L\condpar{x}{\theta, \tau}f\condpar{\theta, \tau}{u, \precision, \alpha, \beta},
\end{equation*}
up to normalization. Thus, by \eqref{eq:likedensity} and \eqref{eq:priordensity} we find the negative-log equality
\begin{align*}
    &-\left(\alpha' - 1 + \frac{\rdim}{2}\right) \Log \pre + \frac{\pre}{2}  \left(\theta - u'\right)^\transpose \precision' \left(\theta - u'\right) + \beta' \pre \\
    &=- \frac{1}{2} \Log \pre + \frac{\pre}{2} \left(x - a^\transpose \theta\right)^2 -\left(\alpha - 1 + \frac{\rdim}{2}\right) \Log \pre + \frac{\pre}{2}  \left(\theta - u\right)^\transpose \precision \left(\theta - u\right) + \beta \pre + (\text{constant term})
\end{align*}
in indeterminates $\theta$ and $\pre$. By expanding and comparing terms we see that
\begin{align*}
    \left(\alpha' - 1 + \frac{\rdim}{2}\right) \Log \pre  &=  \left(\alpha - \frac{1}{2} + \frac{\rdim}{2}\right) \Log \pre, \\
    \frac{\pre}{2} \theta^\transpose \precision' \theta &= \frac{\pre}{2} \theta^\transpose \left(\precision + a a^\transpose \right) \theta.
\end{align*}
Thus, $\alpha'$ and $\precision'$ are found to be $x$-independent with updates
\begin{equation*}
    \alpha' = \alpha + \frac{1}{2}, \quad \precision' = \precision + a a^\transpose
\end{equation*}
as required. Moreover, from
\begin{equation*}
  \frac{\pre}{2} \theta^\transpose \precision' u' = \frac{\pre}{2} \theta^\transpose \left( x a + \precision u \right)  
\end{equation*}
we find the implication $\precision' u' = x a + \precision u$. So from the preceding relation for $\precision'$ we obtain
\begin{equation*}
    u' = \precision'^{-1} \left(x a + \precision u  \right) = \left(\precision + a a^\transpose \right)^{-1} \left(x a + \precision u\right).
\end{equation*}
Finally, a comparison of remaining $\pre$-exclusive terms yields
\begin{equation*}
    \left( \beta' + \frac{1}{2}u'^\transpose \precision' u' \right) \pre = \left( \frac{1}{2} \left(x^2 + u^\transpose \precision u \right) + \beta \right)\pre,
\end{equation*}
where we extract
\begin{equation*}
    \beta' = \beta + \frac{1}{2} \left( x^2 + u^\transpose \precision u- u'^\transpose \precision' u'\right).
\end{equation*}
This completes the proof. 
\end{proof}

We note that \autoref{thm:updates} is given in the form of recurrence relations. The following result shows that under certain assumptions we may choose an initial prior that generates particular simple formulas for iterated parameter values after $n \geq 1$ observations.

\begin{lemma}
\label{thm:updateformula}
Suppose that the context $a \in \mathbb{R}^\rdim$ is normalized. Let $x_1$ be an initial observation. If we put $u_n$, $\precision_n$, $\alpha_n$, $\beta_n$ as prior parameters after $n \geq 1$ observations with initial values $u_1 = x_1 a$, $\precision_1 = I_\rdim$, $\alpha_1 = \frac{1}{2}$. Then given observations $x_1, x_2, \dots, x_n$ we find that
 \begin{align*}
    \precision_n &= I_\rdim + (n-1) a a^\transpose, \\
    u_n &= n \hat{\mu}_n \precision_n^{-1} a, \\
    \alpha_n &= \frac{n}{2}, \\
    \beta_n &= \beta_1 + \frac{1}{2} \sum_{i = 1}^n \left(x_i - \hat{\mu}_n\right)^2,
\end{align*}     
where $\hat{\mu}_n \coloneqq \frac{1}{n} \sum_{i = 1}^n x_i$ is the sample mean. In particular, $a^\transpose u_n = \hat{\mu}_n$ and $a^\transpose \precision_n^{-1} a = 1/n$.
\end{lemma}

\begin{proof}
By \autoref{thm:updates} we have for $n \geq 2$ the following relations:
\begin{align*}
    \precision_n &= \precision_{n-1} + a a^\transpose, \\
    u_n &= \precision_n^{-1} \left(x_n a + \precision_{n-1} u_{n-1}\right), \\
    \alpha_n &= \alpha_{n-1} + \frac{1}{2}, \\
    \beta_n &= \beta_{n-1} + \frac{1}{2} \left( x_n^2 + u_{n-1}^\transpose \precision_{n-1} u_{n-1}- u_n^\transpose \precision_n u_n\right).
\end{align*}
With initial conditions $\precision_1 = I_\rdim$ and $\alpha_1 = \frac{1}{2}$ we clearly have
\begin{align*}
    \precision_n &= I_\rdim + (n-1) a a^\transpose, \\ \alpha_n &= \frac{n}{2},
\end{align*}
for $n \geq 1$ as required. Note that
\begin{equation*}
  u_1 = x_1 a = \hat{\mu}_1 I_\rdim^{-1} a. 
\end{equation*}
So for any $n > 1$, we propose the hypothesis $u_{n-1} = (n-1)\hat{\mu}_{n-1} \precision_{n-1}^{-1} a$. This implies
\begin{align*}
    u_n &= \precision_n^{-1} \left( x_n a + \precision_{n-1} u_{n-1}\right) = \precision_n^{-1} \left( x_n a + (n-1) \hat{\mu}_{n-1} a \right) = n \hat{\mu}_{n} \precision_n^{-1} a.
\end{align*}
Thus by induction $u_n = n \hat{\mu}_{n} \precision_n^{-1} a$ for $n \geq 1$. Finally we proceed with the required statement for $\beta_n$. Put 
\begin{equation*}
   w \coloneqq \precision_n^{-1} a = \left(I_\rdim + (n-1) a a^\transpose \right)^{-1} a
\end{equation*}
and recall that $a^\transpose a = 1$. Then $\left(I_\rdim + (n-1) a a^\transpose \right) w = a$ hence $n a^\transpose w = 1$ and therefore $a^\transpose \precision_n^{-1} a = 1/n$. So using the preceding facts we note
\begin{align*}
    u_n^\transpose \precision_n u_n = n^2 \hat{\mu}_n^2 a^\transpose \precision_n^{-1} \precision_n \precision_n^{-1} a = n^2 \hat{\mu}_n^2 a^\transpose \precision_n^{-1} a = n \hat{\mu}_n^2.
\end{align*}
Thus by telescoping we find
\begin{align*}
    \beta_n &= \beta_{n-1} + \frac{1}{2} \left( x_n^2 + u_{n-1}^\transpose \precision_{n-1} u_{n-1}- u_n^\transpose \precision_n u_n\right) \\
    &=\beta_1 + \sum_{i = 2}^n \frac{1}{2} \left( x_i^2 + u_{i-1}^\transpose \precision_{i-1} u_{i-1}- u_i^\transpose \precision_i u_i\right) \\
    &= \beta_1 + \frac{1}{2} \left( \left(\sum_{i = 2}^n x_i^2\right) + u_{1}^\transpose \precision_{1} u_{1}- u_n^\transpose \precision_n u_n \right) \\
    &= \beta_1 + \frac{1}{2} \left( \left(\sum_{i = 1}^n x_i^2\right) - n \hat{\mu}_n^2 \right) = \beta_1 + \frac{1}{2} \sum_{i = 1}^n \left(x_i - \hat{\mu}_n\right)^2.
\end{align*}
This concludes the proof.
\end{proof}

\subsection{Algorithm with Normal-Gamma Priors}
\begin{figure}[t]
    \centering
    \begin{subfigure}[b]{0.32\textwidth}
        \includegraphics[width=\linewidth]{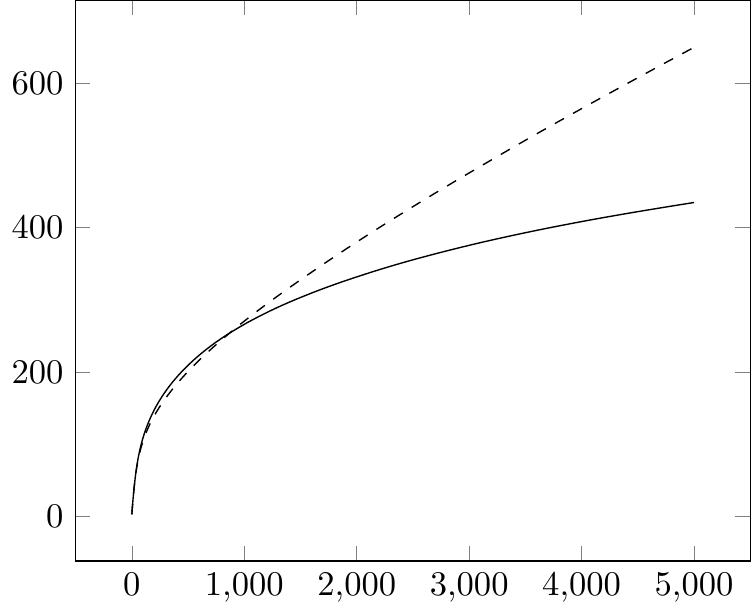}
        \caption{$(\alpha^*, \beta^*) = (3,2)$ \label{fig:ga}}
    \end{subfigure}
    \begin{subfigure}[b]{0.32\textwidth}
        \includegraphics[width=\linewidth]{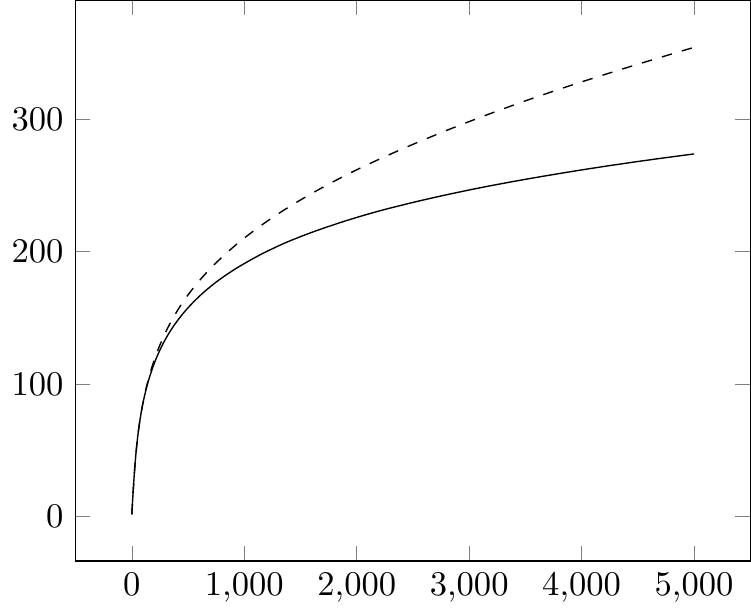}
        \caption{$(\alpha^*, \beta^*) = (3,1)$ \label{fig:gb}}
    \end{subfigure}
    \begin{subfigure}[b]{0.32\textwidth}
        \includegraphics[width=\linewidth]{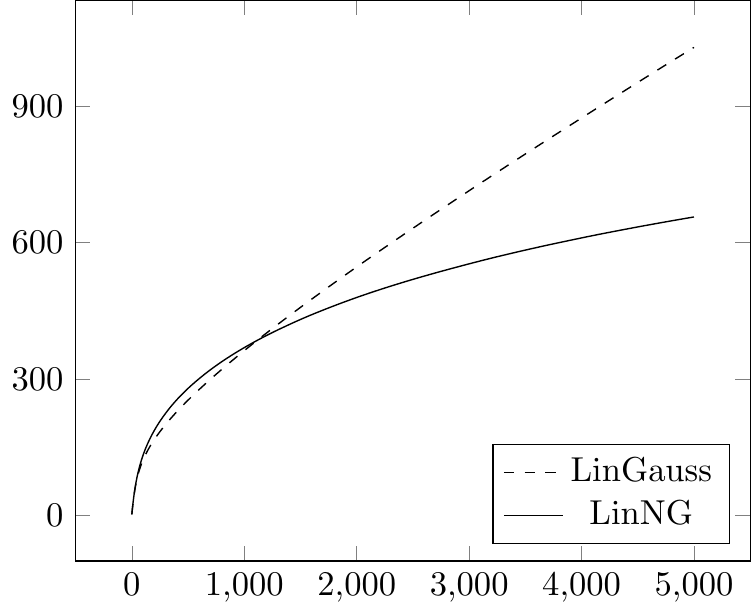}
        \caption{$(\alpha^*, \beta^*) = (3,3)$ \label{fig:gc}}
    \end{subfigure}
    \caption{Estimations of the Bayesian regret given 30 normalized contexts in $\mathbb{R}^5$ and identical priors $\normalgamma(0, I_5, \alpha^*, \beta^*)$ for each arm. Values are computed by averaging simulated regret of \numprint{10000} bandits over \numprint{5000} rounds. In (a) the linear Gaussian TS-algorithm (LinGauss), with assumed fixed variance $\sigma^2=1$, is correctly specified in expectation since $\beta^*/(\alpha^* - 1) = 1$, but almost surely misspecified in each realized case. The degraded performance versus the normal-gamma algorithm (LinNG) is further evident in (b) and (c), where the Gaussian algorithm is also misspecified in expectation ($\sigma^2 = 0.5$ and $\sigma^2 = 1.5$ respectively).}
    \label{fig:regretcomp}
\end{figure}

We note that since the arms are independent, the updates of \autoref{alg:ts} can be separated and simplified.  That is, for each action $k$ we maintain a coherent prior $\normalgamma(u^{(k)}, \precision^{(k)}, \alpha^{(k)}, \beta^{(k)})$ to represent our uncertainty of $(\theta_k, \tau_k) \in \mathbb{R}^d \times \mathbb{R}^+$. Explicitly we have the following algorithm: 
\begin{algorithm}[H]
\caption{Normal-Gamma Linear Bandit TS}\label{alg:ngts}
\begin{algorithmic}
\Require Action set $\mathcal{A}$, distributions $\normalgamma(u^{(k)}, \precision^{(k)}, \alpha^{(k)}, \beta^{(k)})$, $k \in [K]$
\For{$t = 1, 2, \dots T$}
\State For each $k \in [K]$, sample $\Tilde{\tau}_k \sim \Gam(\alpha^{(k)}, \beta^{(k)})$
\State For each $k \in [K]$, sample $\Tilde{\theta}_k \sim \mathcal{N}_d \left(u^{(k)},\left(\Tilde{\tau}_k \precision^{(k)}\right)^{-1}\right)$
\State Choose $A_t \in \argmax_{k \in [K]} a_k^\transpose \Tilde{\theta}_k$
\State Observe $X_t$
\State For $k = A_t$, update $\normalgamma(u^{(k)}, \precision^{(k)}, \alpha^{(k)}, \beta^{(k)})$ with $X_t$ and Proposition~\ref{thm:updates}
\EndFor
\end{algorithmic}
\end{algorithm}
% \todo{Efterföljande text är en del i motiveringen}
\noindent \autoref{fig:regretcomp} shows examples of the degraded performance which may occur if the TS-algorithm is misspecified in its choice of prior. In the experiments a set of 30 normalized contexts in $\mathbb{R}^5$ are chosen uniformly in $[-1/\sqrt{5},1/\sqrt{5}]^5$. The underlying Bayesian prior then dictates that environments should be sampled as $(\theta_k^*, \tau_k^*) \sim \normalgamma(0, I_5, \alpha^*, \beta^*)$, $k \in [30]$. Rewards for each environment and arm is then sampled via $\mathcal{N}(a_k^\transpose \theta_k^*, 1/\tau_k^*)$. The Bayesian regret is estimated by sampling \numprint{10000} bandits per experiment and averaging accumulated regret up to \numprint{5000} rounds. For comparison, a Gaussian linear TS-algorithm is included which assumes fixed precisions $\tau_k^* = 1$, i.e., a fixed variance of $\sigma^2 = 1$ for all rewards. Incidentally, with Gaussian TS-priors $\mathcal{N}_5(u_k, \precision_k)$, this corresponds precisely to using the TS-updates of $u'$ and $\precision'$ in \autoref{thm:updates}. 

In \autoref{fig:ga} environments are sampled with $(\alpha^*, \beta^*) = (3,2)$, which implies that the Gaussian algorithm (LinGauss) is correctly specified in expectation since $\expect{1/\tau_k} = \beta^*/(\alpha^* - 1) = 1$, although we almost surely have a misspecified algorithm in realization which degrades the performance. This means that in most environments the covariance will either converge to 0 too quickly, which implies too little explorative actions, or go to 0 too slowly, which leads to an excess of explorative actions. This is further evident in \autoref{fig:gb} and \autoref{fig:gc}, where we have also have a shift in expectation of $1/\tau_k$ with $(\alpha^*, \beta^*) = (3,1)$ and $(\alpha^*, \beta^*) = (3,3)$ respectively. This implies that on average the Gaussian algorithm should perform excessive exploration in the first case ($\sigma^2=0.5$) or display a tendency to get stuck on non-optimal arms in the latter case ($\sigma^2=1.5$). This issue is removed by the choice of a proper prior, which in this case is the normal-gamma prior (LinNG).
\section{Distributions and Bounds}
\label{sec:distbounds}
In this section we present results concerning distributions of parameters and upper bounds that will be utilized in later analysis. Particularly important is the distribution of the sample mean given a fixed environment and the asymptotic behavior of the expectation of random variables involving the immediate regret and model precisions.
% \todo[color=orange]{Fixed.}\todo{Här behövs lite metatext om avsnittet}

It is clear that there are only two random quantities in \autoref{thm:updateformula} with respect to random observed outcomes and environment uncertainty, namely $u_n$ and $\beta_n$. However, conditioned on model parameters their respective distributions follow immediately by their update formulae.  
\begin{lemma}
\label{thm:conddist}
Let an arm of an instance $\mathcal{E}$ be defined by parameters $(a, \theta, \tau)$ such that observed rewards are drawn iid from $\mathcal{N}\left(a^\transpose \theta, (\tau)^{-1} \right)$. If the context $a \in \mathbb{R}^d$ is normalized and prior parameters $u_n$, $\beta_n$ are updated according to \autoref{thm:updateformula}. Then the following statements hold for a given $\mathcal{E}$:
\begin{enumerate}[(i)]
    \item $a^\transpose u_n = \hat{\mu}_n \sim \mathcal{N}\left(a^\transpose \theta, \left(n \tau\right)^{-1} \right)$ for $n \geq 1$,
    \item $\beta_n \sim \beta_1 + \frac{1}{2 \tau}\chi_{n-1}^2$ for $n \geq 2$,
    \item $a^\transpose u_n$ and $\beta_n$ are conditionally independent for $n \geq 2$.
\end{enumerate}
\end{lemma}

\begin{proof}
From \autoref{thm:updateformula} we recall that $a^\transpose \precision_n^{-1} a = 1/n$. Hence, the first statement is an immediate consequence of
\begin{equation*}
    a^\transpose u_n = n \hat{\mu}_n a^\transpose \precision_n^{-1} a = \hat{\mu}_n
\end{equation*}
and
\begin{equation*}
    \hat{\mu}_n = \frac{1}{n} \sum_{i = 1}^n X_i = \sum_{i = 1}^n \frac{X_i}{n} ,
\end{equation*}
where we have $X_i/n \sim \mathcal{N}\left(a^\transpose \theta/n, 1 / (\tau n^2) \right)$. Thus, by the linear combination of Gaussian variables we obtain
\begin{equation*}
    a^\transpose u_n \sim \mathcal{N}\left( \sum_{i = 1}^n \frac{a^\transpose \theta}{n} , \sum_{i = 1}^n \frac{1}{n^2 \tau} \right) = \mathcal{N}\left( a^\transpose \theta, \frac{1}{n \tau} \right)
\end{equation*}
as required.

The last two statements follow from well-known statistical facts concerning iid samples $(X_i)_{i = 1}^n$ drawn from a normal distribution of mean $\mu$ and variance $\sigma^2$. If we let $\overline{X} = \frac{1}{n} \sum_{i = 1}^n X_i$, then the characteristic function of $\left(\overline{X}, X_i - \overline{X}\right)$ obeys
\begin{equation*}
    \varphi_{\overline{X}, X_i - \overline{X}}(s,t) =\varphi_{\overline{X}}(s)\varphi_{X_i - \overline{X}}(t), 
\end{equation*}
where we have used the fact that $(X_i)_{i = 1}^n$ are Gaussian and independent. Hence $\overline{X}$ is independent of $X_i - \overline{X}$, $i = 1, 2, \dots, n$, and therefore of $\sum_{i = 1}^n \left(X_i - \overline{X}\right)^2$. This implies that $a^\transpose u_n = \hat{\mu}_n$ and $\beta_n = \beta_1 + \frac{1}{2} \sum_{i=1}^n (x_i -  \hat{\mu}_n)^2$ are independent. Moreover, we have $\sum_{i = 1}^n \left(X_i - \overline{X}\right)^2 \sim \sigma^2 \chi_{n-1}^2$ (see for example \textcite{cochran1934distribution}), so with $ \sigma^2 = 1/\tau$ we obtain
\begin{equation*}
   \beta_n \sim \beta_1 + \frac{1}{2 \tau} \chi_{n-1}^2
\end{equation*}
as required.
\end{proof}
We recall from \autoref{sec:slb} that the immediate regret $\Delta_k = \mu^* - \mu_k$ measures the difference between the optimal mean reward and the mean reward for arm $k$. If we apply uncertainty under the Bayesian prior, then it is defined by conditional Gaussian variables. It follows that functional values involving $\Delta_k$ may be unbounded in range but not in expectation. Specifically, if $T_0$ is the smallest sampled precision for an environment, then our next result bounds the expectations of $\Delta_k/T_0^\ell$ for non-negative integers $\ell$ under certain conditions on the prior.
\begin{lemma}
\label{thm:deltabounds}
Let $(X_1, T_1), (X_2, T_2), \dots, (X_K, T_K)$ be a mutually independent sequence drawn from 
\begin{equation*}
   \normalgamma(0, \lambda_k^*, \alpha^*, \beta^*), \ k = 1,2, \dots, K,  
\end{equation*}
such that for each $k$,
\begin{align*}
   X_k \mid T_k &\sim \mathcal{N}\left(0, 1/(\lambda_k^* T_k)\right), \\
   T_k &\sim \Gam(\alpha^*, \beta^*).
\end{align*}
Put $\lambda_0 \coloneqq \min_{k} \lambda_k$ and $T_0 \coloneqq \min_{k} T_k$. If we let $\Delta_k \coloneqq \max_{i} X_i - X_k$, then
\begin{equation*}
    \expect{\Delta_k/T_0^\ell} <\sqrt{\frac{2 \Log K}{\lambda_0}} \expect{\frac{1}{T_0^{\ell + 1/2}}}
\end{equation*}
for nonnegative integers $\ell$.
\end{lemma}

\begin{proof}
Let $\mathfrak{T} \coloneqq \sigma\left(T_1, T_2, \dots, T_k\right)$ be the $\sigma$-algebra generated by $\left(T_k\right)_{k = 1}^K$. If $X_{\max} \coloneqq \max_i X_i$, then
\begin{equation*}
  \condexpect{\Delta_k}{\mathfrak{T}} = \condexpect{X_{\max}}{\mathfrak{T}}.
\end{equation*}
So by Jensen's inequality, given any $\gamma > 0$, we obtain
\begin{align*}
\label{eq:conddeltaineq}
    \exp\left(\gamma \condexpect{\Delta_k}{\mathfrak{T}}\right) &\leq \condexpect{    \exp\left(\gamma X_{\max}\right)}{\mathfrak{T}}
    < \sum_{i=1}^K \condexpect{\Exp{\gamma X_i}}{\mathfrak{T}} \\
    &=\sum_{i = 1}^K \Exp{ \frac{\gamma^2}{2 \lambda_i T_i} } 
    \leq K \Exp{ \frac{\gamma^2}{2 \lambda_0 T_0}}.
\end{align*}
Hence,
\begin{equation*}
    \condexpect{\Delta_k}{\mathfrak{T}} < \frac{\Log K}{\gamma} + \frac{\gamma}{2 \lambda_0 T_0} \quad (\gamma > 0), 
\end{equation*}
where the right hand side is minimized by $\gamma = \sqrt{2 \lambda_0 T_0 \Log K}$. It follows that
\begin{equation*}
    \condexpect{\Delta_k}{\mathfrak{T}} <  2 \sqrt{\frac{\Log K}{2 \lambda_0 T_0}} = \sqrt{\frac{2 \Log K}{ \lambda_0 T_0}}.
\end{equation*}
Thus we obtain 
\begin{align*}
    &\expect{\Delta_k/T_0^\ell} = \expect{\condexpect{\Delta_k/T_0^\ell}{\mathfrak{T}}} 
    < \expect{\sqrt{\frac{2 \Log K}{\lambda_0}} \frac{1}{T_0^{\ell + 1/2}}}
\end{align*}
as required. This concludes the proof.
\end{proof}
We are left with an expectation $\expect{1/T_0^{\ell + 1/2}}$ in \autoref{thm:deltabounds}, and in our analysis we will be concerned about the asymptotic behavior of this expression as the number of arms $K$ grows larger. We note that $1/T_0$ can also be seen as the maximum statistic of an iid sequence of inverse gamma variables $1/T_1, 1/T_2, \dots, 1/T_K$. It turns out that finding bounds on the expectation of powers of this statistic is relatively straight forward by Jensen's inequality. Explicitly we have the following result. 
\begin{lemma}
\label{thm:boinvgamma}
Let $X_1, X_2, \dots, X_K$ be an iid sequence of $\invgamma(\alpha,\beta)$ random variables with $\alpha > 1$. Put $X_{(K)} \coloneqq \max_k X_k$. If a fixed $\gamma$ satisfies $1 < \gamma < \alpha$, then
\begin{equation*}
    \expect{X_{(K)}^{p}} = \mathcal{O}\left(K^{p / \gamma}\right)
\end{equation*}
for any $p < \gamma$.
\end{lemma}
\begin{proof} Recall that for $X \sim \invgamma(\alpha, \beta)$ we have
\begin{equation*}
    \expect{X^\gamma} = \frac{\beta^\gamma \Gamma(\alpha - \gamma)}{\Gamma(\alpha)}  < \infty
\end{equation*}
whenever $\gamma < \alpha$. Further, note that if $1 < \gamma < \alpha$ and $p < \gamma$ then the function $\phi(x) \coloneqq x^{\gamma/p}$ defined on the positive reals is convex and strictly increasing. Thus by Jensen's inequality we obtain
\begin{align*}
    &\expect{X_{(K)}^p}^{\gamma / p} = \phi\left(\expect{X_{(K)}^p}\right) \leq \expect{\phi \left(X_{(K)}^p\right)} = \expect{X_{(K)}^\gamma}\\
    &\leq \expect{\sum_{k=1}^K X_k^\gamma} = K \frac{\beta^\gamma \Gamma(\alpha - \gamma)}{\Gamma(\alpha)},
\end{align*}
hence $\expect{X_{(K)}^{p}} = \mathcal{O}\left(K^{p / \gamma}\right)$ as required.
\end{proof}
We conclude this section with a useful concentration bound, which concerns $\beta_n$ in \autoref{thm:updateformula}. Namely, for any $\chi^2$-distribution we have the following (see \textcite[\S 4.1, Lemma~1]{laurent2000adaptive}).
\begin{lemma}
\label{thm:laurent}
Let $U$ be a $\chi^2$ statistic with $D$ degrees of freedom. Then for any positive $x$,
\begin{equation*}
    \prob{U - D \geq 2 \sqrt{D x} + 2 x} \leq \exp(-x).
\end{equation*}
\end{lemma}

\section{Regret Analysis}
\label{sec:regretanalysis}
In this section we will try to bound the regret of \autoref{alg:ngts} under the assumptions of \autoref{thm:updateformula} and ultimately prove the \hyperlink{thm:mainlemma}{Main Lemma} and \autoref{thm:main}. The technique in the analysis amounts to putting a bound on the random variable $\Regret(T, \theta) = \sum_{k \in K} \condexpect{n_k(T)}{\theta} \Delta_k$ in \eqref{eq:regret} and from there derive a more general bound for the Bayesian regret. 

We can summarize the overall strategy of this section as follows. For a fixed arm and environment we first derive a bound on $\condexpect{n_k(T)}{\theta}$ expressed as two sums of expectations using probabilities $p_k$ and $q_k$ for specific threshold events (\autoref{thm:count}). We then further bound these sums with the aid of a common quantity $\rho_k \in (0,1)$ (\autoref{thm:countrho}). By studying the properties of derived quantities and observing an increasing sequence in $T$ we may thus put an upper bound on the expected sample count in terms of the immediate regret $\Delta_k$ which holds for all $T \geq 2$ (\autoref{sec:mainlemmaproof}). This result is stated in the \hyperlink{thm:mainlemma}{Main Lemma}. Further study of environment specific constants under the Bayesian prior yields our main result in \autoref{sec:mainresultproof}.

\subsection{Definitions and Prior Assumptions}
To save notation we assume that all probabilities and expectations until the last subsection are conditional on model parameters, whereas more specific conditionals, such as those on sampled precision values or events, will be explicitly stated. Moreover, in the analysis, we assume for each bandit a unique optimal arm since the presence of more will only decrease regret. We further assume, in accordance with the initial values of  \autoref{thm:updateformula}, that each arm has been sampled once before the first iteration.

Let the normal-gamma bandit environment $\mathcal{E}$ be defined by true model parameters $(\theta_k, \tau_k)_{k=1}^K$. We recall from \autoref{sec:slb} that conditioned on model parameters we have $\Delta_k \coloneqq \mu^* - \mu_k$ as the expected immediate regret of choosing action $a_k$ over the optimal action. Without loss of generality, put the optimal action index as $k=1$, with mean reward $\mu_1 = \mu^*$. Let $n_k(t)$ be the random variable for the number of updates of the $k$th arm at the start of episode $t = 1,2, \dots$, where we note that for all $k$, $n_k(1) = 1$ by the initial sampling procedure. 

Without an explicitly defined notation, we assume that each arm $k$ with action context $a$ maintains a distinct set of normal-gamma prior parameters $(u_n^{(k)}, \precision_n^{(k)}, \alpha_n^{(k)}, \beta_n^{(k)})$ for $n = n_k(t)$. By the initial sampling procedure and a starting sample $x$ for each arm, we assume initial values 
\begin{equation*}
    \left(u_1, \precision_1, \alpha_1, \beta_1\right) = \left(x a, I, \frac{1}{2}, 1 \right),
\end{equation*}
such that subsequent updates follow \autoref{thm:updateformula}. 
\subsubsection{Q-values}
In view of \autoref{alg:ngts}, we can see that for each round and arm $k$, the learner samples environment parameters $(\Tilde{\theta}_k, \Tilde{\tau}_k)$ and then makes a decision on the value $Q_k \coloneqq a_k^\transpose \Tilde{\theta}_k$, i.e., an estimated \emph{Q-value} of choosing action $a_k$. Thus, going forward we will use the following property of $Q_k$. 
\begin{lemma}
\label{thm:qvalue}
Let $n \coloneqq n_k(t) \geq 1$ be the number of observed samples at the start of round $t$ for action $a_k$. Put $\hat{\mu}_k$ as the observed sample mean. Given a sample $(\Tilde{\theta}_k, \Tilde{\tau}_k)$ from the prior, if we let $Q_k(t) \coloneqq a_k^\transpose \Tilde{\theta}_k$, then $Q_k(t)$ is conditionally Gaussian and distributed as
\begin{equation*}
    Q_k(t) \sim \mathcal{N}\left(\hat{\mu}_k, \left(n\Tilde{\tau}_k\right)^{-1} \right).
\end{equation*}
\end{lemma}
\begin{proof}
Since $(\Tilde{\theta}_k, \Tilde{\tau}_k) \sim \normalgamma(u_n, \precision_n, \alpha_n, \beta_n)$ we have $\Tilde{\theta}_k \mid  \Tilde{\tau}_k \sim \mathcal{N}_d\left(u_n, \left(\Tilde{\tau}_k \precision_n\right)^{-1}\right)$. Thus by \autoref{thm:updateformula} and well-known properties for weighted sums of multivariate normal components, we obtain
\begin{equation*}
    Q_k(t) \sim \mathcal{N}\left(a_k^\transpose u_n , a_k^\transpose \left(\Tilde{\tau}_k \precision_n\right)^{-1} a_k \right) = \mathcal{N}\left(\hat{\mu}_k, \left(n\Tilde{\tau}_k\right)^{-1} \right)
\end{equation*}
as required.
\end{proof}
\subsubsection{Joint Observation Events}
We now proceed by considering high and low probability events. Put $\beta_n^{(k)}$ as the prior beta parameter of the $k$th arm after $n$ updates, and let $\samplemean_k$ be the corresponding sample mean. Important for us will be to bound probabilities of joint events such as $\set{\samplemean_1 > \mu_1 - \varepsilon, \beta_n^{(1)} \leq \frac{n M }{2}}$ and $\set{\samplemean_k > \mu_k + \varepsilon, \beta_n^{(k)} \leq \frac{n M }{2}}$,
given some carefully chosen positive values for $\varepsilon$ and $M$. To do this we first define governing equations over $k$ that are fundamental to the overall analysis. 

Let $g(x) \coloneqq x + \sqrt{C \Log\left(1 + \frac{x^2}{M}\right)}$ for arbitrary positive constants $C$ and $M$. Then $g(x)$ is continuous and increasing on the positive reals with $g(0) = 0$. Hence for every value $D > 0$ the equation $g(x) = D$ admits a unique solution $x \in (0,D)$. Formally, we make the following explicit definition.
\begin{definition}
\label{def:governing}
Put $\tau_0 \coloneqq \min_{k} \tau_k$. Then for any $M_k > 0$ we define $\probc_k$ to be the unique positive root of the equation
\begin{equation*}
    \probc + \sqrt{\frac{1}{\tau_0}\Log\left(1 + \frac{\probc^2}{M_k}\right)} = \frac{\Delta_k}{2}.
\end{equation*}
Thus, if $\varepsilon_k \coloneqq \sqrt{\frac{1}{\tau_0}\Log\left(1 + \frac{\probc_k^2}{M_k}\right)}$ then $\probc_k + \varepsilon_k = \frac{\Delta_k}{2}$ and
\begin{equation*}
    \rho_k \coloneqq \exp\left(-\frac{\tau_0 \varepsilon_k^2}{2}\right) = \left(1 + \frac{\probc_k^2}{M_k}\right)^{-\frac{1}{2}}.
\end{equation*}
\end{definition}

We recall from \autoref{def:cd} the environment dependent values $\tau_0 \coloneqq \min_k \tau_k$ and 
\begin{equation*}
    C(\tau_0) \coloneqq \frac{2}{\tau_0} \sqrt{\frac{ \tau_0}{2} + \Log 2} + \frac{1}{\tau_0} \left(2 \Log 2 + 1 \right) + 3.
\end{equation*}
By a derivation of the particular form for $C(\tau_0)$, we show the existence of values for $M_k$ which by \autoref{thm:laurent} induces bounds on the tail end concentrations of $\beta_n^{(1)}$, $\beta_n^{(k)}$ in terms of $\rho_k$.
\begin{lemma}
\label{thm:mkthm}
Let $C \coloneqq C(\tau_0)$. If
\begin{equation*}
M_k \coloneqq \left \{ \begin{array}{cc}
     C \Delta_k^2, & \Delta_k > 1, \\
     C & \Delta_k,  \leq 1,
\end{array} \right.
    % M_k \coloneqq 1 + \frac{1}{\tau_0} \left( \sqrt{\tau_0\frac{\Delta_k^2}{2} + \Log 16} + \left(\tau_0 \frac{\Delta_k^2}{4} + \Log 4\right) + 1  \right).
\end{equation*}
then for integers $n \geq 1$ the corresponding induced $\rho_k$ yields
\begin{align*}
    \prob{\beta_n^{(1)} > \frac{n M_k}{2}} &\leq \frac{1}{2}\rho_k^n, \\
    \prob{\beta_n^{(k)} > \frac{n M_k}{2}} &\leq \frac{1}{2}\rho_k^n.
\end{align*}
\end{lemma}
\begin{proof}
We recall the initial value $\beta_1^{(k)} = 1$ for all $k$. Moreover, $C > 3$ so the result trivially holds for the case $n = 1$.

For the case $n \geq 2$, from \autoref{thm:conddist} we know that $\beta_n^{(k)} = 1 + \frac{1}{2 \tau_k} U$, where $U \sim \chi^2_{n-1}$. Put
\begin{equation*}
    g(x) \coloneqq 2 \sqrt{(n-1) x} + 2 x + (n - 1)
\end{equation*}
and recall that $\rho_k = \exp\left(-\frac{\tau_0 \varepsilon_k^2}{2}\right)$ by \autoref{def:governing}. Then by \autoref{thm:laurent},
\begin{equation*}
    \prob{2 \tau_k \left(\beta_n^{(k)} - 1 \right) \geq g\left(\frac{n \tau_0 \varepsilon_k^2}{2} + \Log 2 \right)} \leq \frac{1}{2} \exp\left(-\frac{n \tau_0 \varepsilon_k^2}{2}\right) = \frac{1}{2} \rho_k^n,
\end{equation*}
or equivalently
\begin{equation*}
    \prob{\beta_n^{(k)} \geq 1 + \frac{1}{2 \tau_k} \left( 2 \sqrt{(n-1)\left(\frac{n \tau_0 \varepsilon_k^2}{2} + \Log 2 \right)} + 2 \left(\frac{n \tau_0 \varepsilon_k^2}{2} + \Log 2 \right) + (n-1) \right)} \leq \frac{1}{2} \rho_k^n.
\end{equation*}
Note that since $\varepsilon_k \in (0, \Delta_k/2)$ and $\tau_0 \leq \tau_k$ we have
\begin{align*}
    &\frac{1}{2 \tau_k} \left( 2 \sqrt{(n-1)\left(\frac{n \tau_0 \varepsilon_k^2}{2} + \Log 2 \right)} + 2 \left(\frac{n \tau_0 \varepsilon_k^2}{2} + \Log 2 \right) + (n-1) \right)\\
    &< \frac{1}{2 \tau_0} \left( 2 \sqrt{n\left(\frac{n \tau_0 \Delta_k^2}{2} + \Log 2 \right)} + 2 \left(\frac{n \tau_0 \Delta_k^2}{2} + \Log 2 \right) + n \right) \\
    &=\frac{n}{2} \left( \frac{2}{\tau_0} \sqrt{\frac{ \tau_0 \Delta_k^2}{2} + \frac{\Log 2}{n}} + \frac{2}{\tau_0} \left(\frac{\tau_0 \Delta_k^2}{2} +  \frac{\Log 2}{n} \right) + \frac{1}{\tau_0} \right) \\
    &\leq \frac{n}{2} \left(\Delta_k^2  + \frac{2 }{\tau_0 \Delta_k^2}  \sqrt{\frac{ \tau_0}{2} + \Log 2} + \frac{2}{\tau_0} \Log 2 + \frac{1}{\tau_0} \right). 
\end{align*}
Thus
\begin{equation}
\label{eq:mkvalue}
        \prob{\beta_n^{(k)} > \frac{n}{2} \left(\Delta_k^2  + \frac{2}{\tau_0} \sqrt{\frac{ \tau_0 \Delta_k^2}{2} + \Log 2}  + \frac{2}{\tau_0} \Log 2 + \frac{1}{\tau_0} + 2 \right)} \leq \frac{1}{2} \rho_k^n.
\end{equation}
So we let 
\begin{equation*}
    C \coloneqq \frac{2}{\tau_0} \sqrt{\frac{ \tau_0}{2} + \Log 2} + \frac{1}{\tau_0} \left(2 \Log 2 + 1 \right) + 3.
\end{equation*}
From \eqref{eq:mkvalue} it is then clear that if $\Delta_k > 1$ then
\begin{align*}
    &\Delta_k^2  + \frac{2}{\tau_0} \sqrt{\frac{ \tau_0 \Delta_k^2}{2} + \Log 2}  + \frac{2}{\tau_0} \Log 2 + \frac{1}{\tau_0} + 2  \\
    &<\frac{2}{\tau_0} \sqrt{\frac{ \tau_0 \Delta_k^4}{2} + \Log 2 \Delta_k^4}  + \Delta_k^2 \left(\frac{2}{\tau_0} \Log 2 + \frac{1}{\tau_0} + 3\right) = C \Delta_k^2 = M_k.
\end{align*}
Conversely if $\Delta_k \leq 1$ then 
\begin{equation*}
        \Delta_k^2  + \frac{2}{\tau_0} \sqrt{\frac{ \tau_0 \Delta_k^2}{2} + \Log 2}  + \frac{2}{\tau_0} \Log 2 + \frac{1}{\tau_0} + 2 \leq C = M_k.
\end{equation*}
Thus in either case we obtain $$\prob{\beta_n^{(k)} > \frac{n M_k}{2}} \leq \frac{1}{2}\rho_k^n$$ as required. We note that the same derivation holds for $\beta_n^{(1)}$ by the fact that $\tau_0 \leq \tau_1$. This concludes the proof.
\end{proof}
Combining \autoref{thm:mkthm} with the conditional Gaussian properties of sample means we may thus derive the following bounds.
\begin{lemma}
\label{thm:jointevent}
If $M_k$ is defined by \autoref{thm:mkthm}, then for integers $n \geq 1$,
\begin{align*}
    \prob{\samplemean_1 > \mu_1 - \varepsilon_k , \beta_n^{(1)} \leq \frac{n M_k}{2}} &\geq 1 - \rho_k^n, \\
        \prob{\set{\samplemean_k > \mu_k + \varepsilon_k} \cup \set{ \beta_n^{(k)} > \frac{n M_k}{2}}} &\leq \rho_k^n.
\end{align*}
\end{lemma}
\begin{proof}
From the first statement of \autoref{thm:conddist} we know that $\samplemean_k \sim \mathcal{N}\left(\mu_k, 1/(n \tau_k) \right)$ after $n$ observed samples. Thus using \autoref{thm:gaussbounds} and the fact that $\tau_0 = \min\set{\tau_1, \tau_k}$ we have
\begin{align*}
    \prob{\samplemean_1 > \mu_1 - \varepsilon_k} &= \prob{\samplemean_1 > \mu_1 - \frac{1}{\sqrt{n \tau_1}}\sqrt{n \tau_1 }\varepsilon_k } \geq 1 - \frac{1}{2} \exp \left(\frac{n \tau_1 \varepsilon_k^2}{2}\right) \\
    &\geq 1 - \frac{1}{2} \exp \left(\frac{n \tau_0 \varepsilon_k^2}{2}\right)
    =1 - \frac{1}{2} \rho_k^n.
\end{align*}
Similarly,
\begin{align*}
    \prob{\samplemean_k > \mu_k + \varepsilon_k} = \mathcal{Q} \left(\sqrt{n \tau_k }\varepsilon_k \right) \leq \frac{1}{2} \exp \left(\frac{n \tau_k \varepsilon_k^2}{2}\right) \leq \frac{1}{2} \exp \left(\frac{n \tau_0 \varepsilon_k^2}{2}\right) = \frac{1}{2} \rho_k^n. 
\end{align*}
Thus for the case $n = 1$, the result holds trivially since $\prob{\beta_1^{(1)} \leq \frac{M_k}{2}} = 1$ and $\prob{ \beta_1^{(k)} > \frac{M_k}{2}} = 0$. 

Moreover for $n \geq 2$, by \autoref{thm:conddist} we also know that $\samplemean_k$ and $\beta_n^{(k)}$ are conditionally independent. 
So with implicit conditional probability of model parameters together with \autoref{thm:mkthm}, we obtain
\begin{align*}
    &\prob{\samplemean_1 > \mu_1 - \varepsilon_k , \beta_n^{(1)} \leq \frac{n M_k}{2}} = \prob{\samplemean_k > \mu_k + \varepsilon_k} \prob{\beta_n^{(1)} \leq \frac{n M_k}{2}} \\
    &\geq \left(1 - \frac{1}{2} \rho_k^n \right)^2 \geq 1 - \rho_k^n.
\end{align*} 
In the same vein,
\begin{align*}
        \prob{\set{\samplemean_k > \mu_k + \varepsilon_k} \cup \set{ \beta_n^{(k)} > \frac{n M_k}{2}}} \leq \frac{1}{2} \rho_k^n  + \frac{1}{2} \rho_k^n = \rho_k^n.
\end{align*}
This completes the proof.
\end{proof}

\subsubsection{Threshold Events}
Fix for non-optimal actions the midway thresholds
\begin{equation}
\label{eq:bk}
b_k \coloneqq \mu_k + \frac{\Delta_k}{2} = \mu^* - \frac{\Delta_k}{2},
\end{equation}
satisfying $\mu_k < b_k < \mu^*$, $k =2,\dots, K$. To simplify notation, we put $B_k(t) \colon Q_k(t) \leq b_k$ as the event of arm $k$ sampling a Q-value less than or equal to $b_k$ at time $t$.
\begin{definition}
\label{def:pkqk}
Given an episode time $t = 1, 2, \dots$ we define
\begin{align*}
    p_k(t) &\coloneqq \prob{Q_1(t) > b_k}, \\
    q_k(t) &\coloneqq \prob{Q_k(t) > b_k} = \prob{B_k^\complement(t)}
\end{align*}
as the probabilities of the optimal arm and arm $k$ sampling higher than $b_k$. Note that $p_k(t) > 0$ since regardless of history $Q_1(t)$ has a continuous distribution. 
\end{definition}

A useful result regarding $B_k(t)$ is by \textcite{agrawal}, which we reformulate and reprove here for clarity.
\begin{lemma}
\label{thm:agone}
At any given time step $t\geq 1$ we have the inequality
\begin{equation*}
    \prob{A_t = k, B_k(t)} \leq \frac{(1 - p_k(t))}{p_k(t)} \prob{A_t = 1, B_k(t)}.
\end{equation*}
\end{lemma}
\begin{proof}
We recall that $B_k(t)$ is the event $Q_k(t) \leq b_k$. So given $B_k \coloneqq B_k(t)$, variables $Q_k \coloneqq Q_k(t)$ and the mutual independence of arms we find
\begin{align*}
  \prob{A_t = k \mid B_k}  &= \prob{\forall i \ Q_i \leq Q_k \mid B_k} \\
  &\leq \prob{\forall i \ Q_i \leq b_k \mid B_k} \\
  &= \prob{Q_1 \leq b_k \mid \forall i \geq 2 \ Q_i \leq b_k, B_k } \prob{\forall i\geq2 \ Q_i \leq b_k \mid B_k} \\
  &= \prob{Q_1 \leq b_k} D \\
  &= \left(1-p_k(t)\right) D,
\end{align*}
where $D \coloneqq \prob{\forall i\geq2 \ Q_i(t) \leq b_k \mid B_k}$. Moreover, 
\begin{align*}
  \prob{A_t = 1 \mid B_k}  &= \prob{\forall i \ Q_i \leq Q_1 \mid B_k} \\
  &\geq \prob{Q_1 > b_k, \ \forall i\geq2 \ Q_i \leq b_k \mid B_k} \\ 
  &= \prob{Q_1 > b_k \mid \forall i \geq 2 \ Q_i \leq b_k, B_k} \prob{\forall i \geq 2 \ Q_i \leq b_k \mid B_k} \\
  &= \prob{Q_1 > b_k} D \\
  &= p_k(t) D,
\end{align*}
which implies $D \leq \prob{A_t = 1 \mid B_k} / p_k(t)$. Thus
\begin{equation*}
    \prob{A_t = k \mid B_k} \leq \frac{(1 - p_k(t))}{p_k(t)} \prob{A_t = 1 \mid B_k}.
\end{equation*}
So by the definition of conditional probability we obtain
\begin{equation*}
    \prob{A_t = k, B_k(t)} \leq \frac{(1 - p_k(t))}{p_k(t)} \prob{A_t = 1, B_k(t)}
\end{equation*}
as required.
\end{proof}

\subsection{Expected Sample Counts}
We know that the expected regret in \eqref{eq:regret} can be formulated as a sum of expected regret over arms, where each term is the product of the expected sample count after $T$ rounds and the immediate regret. So given a fixed environment, we thus have terms of the form $\expect{n_k(T)} \Delta_k$. Our overall goal will be to find a bound on $\expect{n_k(T)}$ expressed as a function of $\Delta_k$. An important step to achieve this goal will be to use the model dependent constant $\rho_k$ in \autoref{thm:mkthm} and then find a function $f$ such that $\expect{n_k(T)} < f(\rho_k)$. Further analysis of both $\rho_k$ and the function will then produce the desired result. 
Thus going forward we will assume that the values of $\rho_k$ is induced by the $M_k$ values in \autoref{thm:mkthm}. Our starting point is the following result, which puts a bound on $\expect{n_k(T)}$ in terms of threshold events $p_k$ and $q_k$.
\begin{lemma}
\label{thm:count}
If we let $s_m$, $r_m$ denote respective random times for which the optimal arm and arm $k$ is sampled for the $m$th time. Then the expected count $\expect{n_k(T)}$ for the $k$th arm after $T \geq 2$ episodes is bounded by
\begin{equation*}
    \expect{n_k(T)} \leq 1 + \sum_{m = 2}^T \expect{\frac{(1 - p_k(s_m))}{p_k(s_m)}} + \sum_{m = 2}^T \expect{q_k(r_m)}.
\end{equation*}
\end{lemma}

\begin{proof}
With initial sampling we know that $n_k(1) = 1$ for all $k$ such that
\begin{equation*}
    \expect{n_k(T)} = 1 + \sum_{t = 2}^T \prob{A_t = k}.
\end{equation*}
Moreover, 
by \autoref{thm:agone} we find 
\begin{align*}
    \sum_{t = 2}^T \prob{A_t = k} &= \sum_{t = 2}^T \prob{A_t = k, B_k(t)} + \prob{A_t = k, B_k(t)^\complement} \\
    &\leq \sum_{t = 2}^T \frac{\left(1 - p_k(t)\right)}{p_k(t)} \prob{A_t = 1} + \sum_{t = 2}^T \prob{A_t = k, B_k(t)^\complement} \\
    &= \sum_{t = 2}^T \frac{\left(1 - p_k(t)\right)}{p_k(t)} \expect{ \mathbbm{1}_{A_t = 1} } + \sum_{t = 2}^T \expect{\mathbbm{1}_{A_t = k} \cdot \mathbbm{1}_{B_k(t)^\complement}} \\
    &\leq \expect{\sum_{t = 2}^{s_T} \frac{\left(1 - p_k(t)\right)}{p_k(t)}\mathbbm{1}_{A_t = 1} } + \expect{\sum_{t = 2}^{r_T} \mathbbm{1}_{A_t = k} \cdot \mathbbm{1}_{B_k(t)^\complement}} \\ 
    &= \sum_{m = 2}^T \expect{\frac{(1 - p_k(s_m))}{p_k(s_m)}} + \sum_{m = 2}^T \expect{q_k(r_m)}.
\end{align*}
which proves the result.
\end{proof}

Our next two results involves bounds on the individual terms $\expect{\frac{(1 - p_k(s_m))}{p_k(s_m)}}$ and $\expect{q_k(r_m)}$  in \autoref{thm:count}. Specifically we aim to express these bounds using the environment dependent quantity $\rho_k$ of \autoref{def:governing}, with values induced by \autoref{thm:mkthm}.
\begin{lemma}
\label{thm:pk}
If $m$ is an integer greater than or equal to $2$ and $n = m - 1$, then
\begin{equation*}
    \expect{\frac{(1 - p_k(s_m))}{p_k(s_m)}} < \rho_k^n \left(\frac{2}{1 - \rho_k^n} - \frac{1}{2-\rho_k^n}\right). 
\end{equation*}
\end{lemma}

\begin{proof}
By definition we have $n = m - 1$ samples digested by the prior $\normalgamma\left(u_n^{(1)}, \precision_n^{(1)}, \alpha_n^{(1)}, \beta_n^{(1)}\right)$ of arm 1 at time $s_m$. Put $\alpha_n \coloneqq \alpha_n^{(1)}$ and $\beta_n \coloneqq \beta_n^{(1)}$. We recall from \autoref{def:pkqk} that
\begin{equation*}
    p_k(t) = \prob{Q_1(t) > b_k},
\end{equation*}
where $b_k$ is the threshold in \eqref{eq:bk}. In addition, by \autoref{thm:qvalue}, if $(\Tilde{\theta}, \Tilde{\tau})$ is sampled from the prior then 
\begin{equation*}
   Q_1(s_m) \sim \mathcal{N}\left(\hat{\mu}_1, \left(n\Tilde{\tau}\right)^{-1} \right). 
\end{equation*}
Since $\rho_k$ is induced by \autoref{thm:mkthm} we have quantities $M_k$, $\probc_k$ and $\varepsilon_k$, where
\begin{align*}
&\probc_k + \varepsilon_k = \frac{\Delta_k}{2}, \\
    &\rho_k = \left(1 + \frac{\probc_k^2}{M_k}\right)^{-1/2}.
\end{align*}
Moreover, by \autoref{thm:jointevent} we also have 
\begin{equation}
\label{eq:pkjoint}
        \prob{\samplemean_1 > \mu_1 - \varepsilon_k , \beta_n \leq \frac{n M_k}{2}} \geq 1 - \rho_k^n
\end{equation}
for the high probability event $\set{\samplemean_1 > \mu_1 - \varepsilon_k , \beta_n \leq \frac{n M_k}{2}}$. 

Assume now that $\set{\samplemean_1 > \mu_1 - \varepsilon_k , \beta_n \leq \frac{n M_k}{2}}$ holds. Then we find that
\begin{align*}
    \samplemean_1 - \probc_k > \mu_1 - \probc_k - \varepsilon_k = \mu_1 - \frac{\Delta_k}{2} = b_k,
\end{align*}
i.e., $b_k$ is upper bounded by $\samplemean_1 - \probc_k$. So by using the Gaussian properties in \autoref{thm:gaussbounds}, we can put a lower bound on the conditional probability $\condprob{Q_1(s_m) > b_k}{\Tilde{\tau}, \samplemean_1 > \mu_1 - \varepsilon_k , \beta_n \leq \frac{n M_k}{2}}$ by
\begin{align*}
    &\condprob{Q_1(s_m) > b_k}{\Tilde{\tau}, \samplemean_1 > \mu_1 - \varepsilon_k , \beta_n \leq \frac{n M_k}{2}} \\
    &> \condprob{Q_1(\tau_n) > \samplemean_1 - \frac{\sqrt{n \Tilde{\tau}}\probc_k}{\sqrt{n \Tilde{\tau}}}}{\Tilde{\tau}, \samplemean_1 > \mu_1 - \varepsilon_k , \beta_n \leq \frac{n M_k}{2}} \\
    &\geq 1 - \frac{1}{2}\exp\left(- \frac{\probc_k^2 n \Tilde{\tau}}{2}\right).
\end{align*}
Since $\alpha_n = \frac{n}{2}$ is deterministic by \autoref{thm:updateformula} and $\beta_n \leq \frac{n M_k}{2}$ by the conditional event, we may marginalize out $\Tilde{\tau}$ and obtain a bound in $\rho_k$:
\begin{align*}
    &\condprob{Q_1(s_m) > b_k}{\samplemean_1 > \mu_1 - \varepsilon_k , \beta_n \leq \frac{n M_k}{2}} \\
    &> \frac{\beta_n^{\alpha_n}}{\Gamma(\alpha_n)} \int_{0}^\infty \left(1 - \frac{1}{2}\exp\left(- \frac{\probc_k^2 n \Tilde{\tau}}{2}\right)\right)  \Tilde{\tau}^{\alpha_n - 1} \exp(- \beta_n \Tilde{\tau}) \ d \Tilde{\tau} \\
    &=1 - \frac{1}{2} \frac{\beta_n^{\alpha_n}}{\Gamma(\alpha_n)} \int_{0}^\infty \Tilde{\tau}^{\alpha_n - 1} \exp\left(- \left(\beta_n + \frac{\probc_k^2 n }{2}\right) \Tilde{\tau}\right) \ d \Tilde{\tau} = 1 - \frac{1}{2}\left(1 + \frac{\probc_k^2 n}{2 \beta_n}\right)^{-\alpha_n} \\
    &\geq 1 - \frac{1}{2} \left(1 + \frac{\probc_k^2 n}{ 2 \frac{n M_k}{2}}\right)^{- \frac{n}{2}} = 1 - \frac{1}{2} \left(1 + \frac{\probc_k^2}{M_k}\right)^{- \frac{n}{2}} = 1 - \frac{1}{2} \rho_k^n.
\end{align*}
Combined with \eqref{eq:pkjoint} this implies that we can lower bound $p_k(s_m)$ with
\begin{align*}
    &p_k(s_m) = \prob{Q_1(s_m) > b_k} \\ 
    &\geq \condprob{Q_1(s_m) > b_k}{\samplemean_1 > \mu_1 - \varepsilon_k , \beta_n \leq \frac{n M_k}{2}} \prob{\samplemean_1 > \mu_1 - \varepsilon_k , \beta_n \leq \frac{n M_k}{2}}\\
    & > \left( 1 - \frac{1}{2} \rho_k^n\right)\left(1 - \rho_k^n\right).
\end{align*}
Therefore we have
\begin{equation*}
    \expect{\frac{1 - p_k(s_m)}{p_k(s_m)}} = \expect{\frac{1}{p_k(s_m)} - 1 } < \frac{2}{\left(2 - \rho_k^n\right)\left(1 - \rho_k^n\right)} - 1 = \rho_k^n \left(\frac{2}{1 - \rho_k^n} - \frac{1}{2-\rho_k^n}\right)
\end{equation*}
as required.
\end{proof}
\begin{lemma}
\label{thm:qk}
If $m$ is an integer greater than or equal to $2$ and $n = m - 1$, then
\begin{equation*}
    \expect{q_k(r_m)} < \frac{3}{2}\rho_k^n. 
\end{equation*}
\end{lemma}
\begin{proof}
We proceed much in the same way as in the proof of \autoref{thm:pk}. We have $n = m - 1$ samples digested by the prior $\normalgamma\left(u_n^{(k)}, \precision_n^{(k)}, \alpha_n^{(k)}, \beta_n^{(k)}\right)$ of arm $k$ at time $r_m$. We put $\alpha_n \coloneqq \alpha_n^{(k)}$, $\beta_n \coloneqq \beta_n^{(k)}$ and recall from \autoref{def:pkqk} that $q_k(t) = \prob{Q_k(t) > b_k}$. In addition, by \autoref{thm:qvalue}, if $(\Tilde{\theta}, \Tilde{\tau})$ is sampled from the prior then $Q_k(r_m)$ is $\mathcal{N}\left(\hat{\mu}_k, \left(n\Tilde{\tau}\right)^{-1} \right)$-distributed. Since $\rho_k$ is induced by \autoref{thm:mkthm} we have quantities $M_k$, $\probc_k$ and $\varepsilon_k$, where
\begin{align*}
&\probc_k + \varepsilon_k = \frac{\Delta_k}{2}, \\
    &\rho_k = \left(1 + \frac{\probc_k^2}{M_k}\right)^{-1/2}.
\end{align*}
Finally, by \autoref{thm:jointevent} we also have 
\begin{equation}
\label{eq:qkjoint}
        \prob{\set{\samplemean_k > \mu_k + \varepsilon_k} \cup \set{ \beta_n^{(k)} > \frac{n M_k}{2}}} \leq \rho_k^n.
\end{equation}
However, assume instead that the event $\set{\samplemean_k \leq \mu_k + \varepsilon_k , \beta_n \leq \frac{n M_k}{2}}$ holds. Then
\begin{equation*}
    \samplemean_k + \probc_k \leq \mu_k + \varepsilon_k + \probc_k = \mu_k + \frac{\Delta_k}{2} = b_k,
\end{equation*}
i.e., $b_k$ is lower bounded by $\samplemean_k + \probc_k$.
So by \autoref{thm:gaussbounds} we obtain the bound
\begin{align*}
    &\condprob{Q_k(r_m) > b_k}{\Tilde{\tau}, \samplemean_k \leq \mu_k + \varepsilon_k , \beta_n \leq \frac{n M_k}{2}} \\ 
    &\leq \condprob{Q_k(r_m) > \samplemean_k +  \frac{\probc_k\sqrt{n \Tilde{\tau}}  }{\sqrt{n \Tilde{\tau}}}}{\Tilde{\tau}, \samplemean_k \leq \mu_k + \varepsilon_k , \beta_n \leq \frac{n M_k}{2}} \\
    &= \mathcal{Q}( \probc_k \sqrt{n \Tilde{\tau}}) \leq \frac{1}{2}\exp\left(- \frac{\probc_k^2 n \Tilde{\tau}}{2}\right).
\end{align*}
We recall that $\alpha_n = \frac{n}{2}$ is deterministic and that $\beta_n$ is upper bounded by $\frac{n M_k }{2}$ given the conditional event. So by the same marginalization as in \autoref{thm:pk} this implies
\begin{align}
    &\condprob{Q_k(r_m) > b_k}{\samplemean_k \leq \mu_k + \varepsilon_k , \beta_n \leq \frac{n M_k}{2}} 
    \leq \frac{1}{2}\left(1 + \frac{\probc_k^2 n}{2 \beta_n}\right)^{-\alpha_n} \nonumber \\
    &\leq \frac{1}{2} \left(1 + \frac{\probc^2}{M_k}\right)^{- \frac{n}{2}} = \frac{1}{2} \rho_k^n. \label{eq:condqk}
\end{align}
Thus, with \eqref{eq:qkjoint} and \eqref{eq:condqk} we may extract an upper bound on $q_k(r_m)$ by
\begin{align*}
&\prob{Q_k(r_m) > b_k} \\
&= \condprob{Q_k(r_m) > b_k}{\set{\samplemean_k > \mu_k + \varepsilon_k}\cup \set{\beta_n > \frac{n M_k}{2}} }\prob{\set{\samplemean_k > \mu_k + \varepsilon_k}\cup \set{\beta_n > \frac{n M_k}{2}}} \\
&\quad + \condprob{Q_k(r_m) > b_k}{\samplemean_k \leq \mu_k + \varepsilon_k, \beta_n \leq \frac{n M_k}{2}} \prob{\samplemean_k \leq \mu_k + \varepsilon_k, \beta_n \leq \frac{n M_k}{2}} \\
&<\prob{\set{\samplemean_k > \mu_k + \varepsilon_k}\cup \set{\beta_n > \frac{n M_k}{2}}} + \condprob{Q_k(r_m) > b_k}{\samplemean_k \leq \mu_k + \varepsilon_k, \beta_n \leq \frac{n M_k}{2}} \\
&\leq \frac{1}{2} \rho_k^n + \rho_k^n = \frac{3}{2} \rho_k^n. 
\end{align*}
This completes the proof.
\end{proof}

Armed with the bounds of \autoref{thm:pk} and \autoref{thm:qk} on the individual terms $ \expect{\frac{(1 - p_k(s_m))}{p_k(s_m)}}$ and $\expect{q_k(r_m)}$ of \autoref{thm:count}, we may now put an upper bound on the expected count $\expect{n_k(T)}$ in terms of the quantity $\rho_k$ induced by \autoref{thm:mkthm}. Specifically, we have the following result.
\begin{lemma}
\label{thm:countrho}
The expected sample count is bounded as
\begin{equation*}
   \expect{n_k(T)} < 1 + \rho_k \left(\frac{2}{1 - \rho_k} - \frac{1}{2-\rho_k}\right) + \frac{1}{\Log(1/\rho_k)} \Log \left( \frac{(2 - \rho_k)(1 - \rho_k^{T-1})^2}{(1 - \rho_k)^2(2 - \rho_k^{T-1})}\right) +  \frac{3 \rho_k(1 - \rho_k^{T-1})}{2(1 - \rho_k)}
\end{equation*}
for $T \geq 2$.
\end{lemma}

\begin{proof}
We recall from \autoref{thm:count} that 
\begin{equation*}
    \expect{n_k(T)} \leq 1 + \sum_{m = 2}^T \expect{\frac{(1 - p_k(s_m))}{p_k(s_m)}} + \sum_{m = 2}^T \expect{q_k(r_m)}.
\end{equation*}
Moreover, by \autoref{thm:pk} and \autoref{thm:qk} we have
\begin{align*}
        \expect{\frac{(1 - p_k(s_m))}{p_k(s_m)}} &< \rho_k^n \left(\frac{2}{1 - \rho_k^n} - \frac{1}{2-\rho_k^n}\right), \\ \expect{q_k(r_m)} &< \frac{3}{2}\rho_k^n,
\end{align*}
where $n = m - 1$ is the number of observed samples at decision time $s_m$. Thus
\begin{equation}
\label{eq:sumrho}
    \sum_{m = 2}^T \expect{\frac{(1 - p_k(s_m))}{p_k(s_m)}} + \sum_{m = 2}^T \expect{q_k(r_m)} < \sum_{n = 1}^{T-1} \rho_k^n \left(\frac{2}{1 - \rho_k^n} - \frac{1}{2-\rho_k^n}\right) + \sum_{n = 1}^{T-1} \frac{3}{2} \rho_k^n.
\end{equation}
We note that 
\begin{equation*}
   \rho_k^x \left(\frac{2}{1 - \rho_k^x} - \frac{1}{2-\rho_k^x}\right), 
\end{equation*}
when seen as a function of $x$, is strictly decreasing and positive on $\mathbb{R}^+$. Hence
\begin{gather}
    \sum_{n = 1}^{T-1} \rho_k^n \left(\frac{2}{1 - \rho_k^n} - \frac{1}{2-\rho_k^n}\right) \leq \rho_k \left(\frac{2}{1 - \rho_k} - \frac{1}{2-\rho_k}\right)  + \sum_{n = 2}^{T-1} \int_{n-1}^{n} \rho_k^x \left(\frac{2}{1 - \rho_k^x} - \frac{1}{2-\rho_k^x}\right) \ d x \nonumber \\
    = \rho_k \left(\frac{2}{1 - \rho_k} - \frac{1}{2-\rho_k}\right)  + \int_1^{T-1} \rho_k^x \left(\frac{2}{1 - \rho_k^x} - \frac{1}{2-\rho_k^x}\right) \ d x \nonumber \\
    = \rho_k \left(\frac{2}{1 - \rho_k} - \frac{1}{2-\rho_k}\right) + \frac{1}{\Log(1/\rho_k)} \int_{\rho_k^{T-1}}^{\rho_k} \frac{2}{1 - w} - \frac{1}{2- w} \ d w \nonumber \\
    = \rho_k \left(\frac{2}{1 - \rho_k} - \frac{1}{2-\rho_k}\right) + \frac{1}{\Log(1/\rho_k)} \Log \left( \frac{(2 - \rho_k)(1 - \rho_k^{T-1})^2}{(1 - \rho_k)^2(2 - \rho_k^{T-1})}\right) \label{eq:middle_b}.
\end{gather}
With \eqref{eq:middle_b} and the fact that
\begin{equation*}
   \sum_{n = 1}^{T-1} \frac{3}{2}\rho_k^n = \frac{3 \rho_k(1 - \rho_k^{T-1})}{2(1 - \rho_k)}, 
\end{equation*}
we can thus further bound the right hand side in \eqref{eq:sumrho}. It follows that
\begin{equation*}
    \expect{n_k(T)} < 1 + \rho_k \left(\frac{2}{1 - \rho_k} - \frac{1}{2-\rho_k}\right) + \frac{1}{\Log(1/\rho_k)} \Log \left( \frac{(2 - \rho_k)(1 - \rho_k^{T-1})^2}{(1 - \rho_k)^2(2 - \rho_k^{T-1})}\right) + \frac{3 \rho_k(1 - \rho_k^{T-1})}{2 (1 - \rho_k)},
\end{equation*}
which is our required inequality. This completes the proof.
\end{proof}
Thus, \autoref{thm:countrho} yields an environment dependent bound on the expected sample count for each arm as a function of an opaque value $\rho_k$. However, by the governing equation in \autoref{def:governing} and the logarithmic properties in \autoref{thm:logprops}, we may under the assumptions of \autoref{thm:mkthm} restrict the functional values of $\rho_k$ in terms of model properties $\Delta_k$ and $\tau_0$. Recall from \autoref{def:cd} that $\tau_0 \coloneqq \min_k \tau_k$ and 
\begin{align*}
    C(\tau_0) &\coloneqq \frac{2}{\tau_0} \sqrt{\frac{ \tau_0}{2} + \Log 2} + \frac{1}{\tau_0} \left(2 \Log 2 + 1 \right) + 3, \\
    D(\tau_0) &\coloneqq \frac{8}{\tau_0}\left(1 + \sqrt{\frac{\tau_0 \left(2 C(\tau_0) + \frac{1}{4} \right)}{2}} \right)^2.
\end{align*}
Then explicitly we have the following result.
\begin{lemma}
\label{thm:dbounds}
If $D \coloneqq D(\tau_0)$ then the following inequalities hold:
\begin{equation*}
    \frac{\rho_k}{1 - \rho_k} \leq \frac{1}{\Log\left(\frac{1}{\rho_k}\right)} < \left \{ \begin{array}{cc}
        D, & \Delta_k > 1 , \\
        D/\Delta_k^2, & \Delta_k \leq 1. 
    \end{array}\right.
\end{equation*}
\end{lemma}
\begin{proof}
We recall from \autoref{def:governing} and \autoref{thm:mkthm} the definition
\begin{equation*}
    \rho_k \coloneqq \exp\left(-\frac{\tau_0 \varepsilon_k^2}{2}\right) = \left(1 + \frac{\probc_k^2}{M_k}\right)^{-\frac{1}{2}},
\end{equation*}
where $\varepsilon_k \coloneqq \sqrt{\frac{1}{\tau_0}\Log\left(1 + \frac{\probc_k^2}{M_k}\right)}$, $\probc_k + \varepsilon_k = \Delta_k/2$, and where 
\begin{equation*}
  M_k \coloneqq \left \{ \begin{array}{cc}
     C \Delta_k^2, & \Delta_k > 1, \\
     C & \Delta_k,  \leq 1,
\end{array} \right.  
\end{equation*}
for $C(\tau_0)$. We can thus rewrite the governing equation $\probc_k + \varepsilon_k = \frac{\Delta_k}{2}$ as
\begin{equation*}
    \probc_k + \sqrt{\frac{2}{\tau_0} \Log \left(\frac{1}{\rho_k}\right)} = \frac{\Delta_k}{2}.
\end{equation*}
So by \autoref{thm:logprops} and the fact that $\probc_k \in (0, \Delta_k/2)$ we obtain
\begin{align*}
    \frac{2}{\tau_0} \Log \left(\frac{1}{\rho_k}\right) = \frac{1}{\tau_0} \Log \left(1 + \frac{\probc_k^2}{M_k}\right) \geq \frac{2}{\tau_0} \frac{\probc_k^2}{2 M_k + \probc_k^2} > \frac{2}{\tau_0} \frac{\probc_k^2}{2 M_k + \frac{\Delta_k^2}{4}},
\end{align*}
hence
\begin{equation*}
    \probc_k < \sqrt{\frac{\tau_0 \left(2 M_k + \frac{\Delta_k^2}{4} \right)}{2}} \sqrt{\frac{2}{\tau_0} \Log \left(\frac{1}{\rho_k}\right)}.
\end{equation*}
From the governing equation we arrive at
\begin{equation}
\label{eq:probcbound}
    \frac{\Delta_k}{2} < \left(1 + \sqrt{\frac{\tau_0 \left(2 M_k + \frac{\Delta_k^2}{4} \right)}{2}} \right) \sqrt{\frac{2}{\tau_0} \Log \left(\frac{1}{\rho_k}\right)}.
\end{equation} 
We now complete this proof by considering the different cases concerning the immediate regret $\Delta_k$.
\begin{description}
\item[(Case $\Delta_k > 1$)] If $\Delta_k > 1$ then $M_k = C \Delta_k^2$. Thus by \eqref{eq:probcbound} we find
\begin{align*}
    \frac{\Delta_k}{2} &< \left(\Delta_k + \sqrt{\frac{\tau_0 \left(2 C \Delta_k^2 + \frac{\Delta_k^2}{4} \right)}{2}} \right) \sqrt{\frac{2}{\tau_0} \Log \left(\frac{1}{\rho_k}\right)} \\
    &= \frac{\Delta_k}{2} 2\left(1 + \sqrt{\frac{\tau_0 \left(2 C  + \frac{1}{4} \right)}{2}} \right) \sqrt{\frac{2}{\tau_0} \Log \left(\frac{1}{\rho_k}\right)} \\
    &= \frac{\Delta_k}{2} 2 \sqrt{\frac{\tau_0 D}{8}} \sqrt{\frac{2}{\tau_0} \Log \left(\frac{1}{\rho_k}\right)}.
\end{align*} 
Hence
\begin{equation*}
    \frac{1}{\Log \left(\frac{1}{\rho_k}\right)} <  D
\end{equation*}
as required. Therefore, by \autoref{thm:logprops} this implies
\begin{equation*}
    \frac{\rho_k}{1 - \rho_k} \leq \frac{1}{\Log \left(\frac{1}{\rho_k}\right)} <  D.
\end{equation*}
\item[(Case $\Delta_k \leq 1$)] If $\Delta_k \leq 1$ then $M_k = C$. Thus by \eqref{eq:probcbound} we find
\begin{equation*}
        \frac{\Delta_k}{2} < \left(1 + \sqrt{\frac{\tau_0 \left(2 C + \frac{1}{4} \right)}{2}} \right) \sqrt{\frac{2}{\tau_0} \Log \left(\frac{1}{\rho_k}\right)} = \sqrt{\frac{\tau_0 D}{8}} \sqrt{\frac{2}{\tau_0} \Log \left(\frac{1}{\rho_k}\right)}.
\end{equation*}
Hence
\begin{equation*}
    \frac{1}{\Log \left(\frac{1}{\rho_k}\right)} < \frac{D}{\Delta_k^2}
\end{equation*}
and similarly to the previous case we therefore obtain
\begin{equation*}
    \frac{\rho_k}{1 - \rho_k} < \frac{D}{\Delta_k^2}.
\end{equation*}
\end{description}
This concludes the proof.
\end{proof}

\subsection{Proof of the Main Lemma}
\label{sec:mainlemmaproof}
Thus given the bound in \autoref{thm:countrho} and the inequalities of \autoref{thm:dbounds}, we may now fully bound the expected count for all time horizons $T \geq 2$ and thereby prove the \hyperlink{thm:mainlemma}{Main Lemma}, which states that the regret for each arm is upper bounded by
\begin{equation*}
    \frac{7 D}{2} \left(\Delta_k + \frac{1}{\Delta_k}\right) + 2D\left(\Log D \Delta_k + \frac{1}{\Delta_k} \Log \left(\frac{D}{\Delta_k^2} \right)\right) + 9 \Delta_k.
\end{equation*}
\begin{proof}
    We note that the bounding expression
    \begin{equation*}
        \rho_k \left(\frac{2}{1 - \rho_k} - \frac{1}{2-\rho_k}\right) + \frac{1}{\Log(1/\rho_k)} \Log \left( \frac{(2 - \rho_k)(1 - \rho_k^{T-1})^2}{(1 - \rho_k)^2(2 - \rho_k^{T-1})}\right) +  \frac{3 \rho_k(1 - \rho_k^{T-1})}{2(1 - \rho_k)} + 1
    \end{equation*}
    in \autoref{thm:countrho} is an increasing function in $T$. Hence for all $T \geq 2$ we have
    \begin{align*}
        \expect{n_k(T)} &< \frac{2 \rho_k}{1 -\rho_k} - \frac{\rho_k}{2-\rho_k} + \frac{3 \rho_k }{2(1-\rho_k)} + \frac{1}{\Log(1/\rho_k)} \Log \left( \frac{2 - \rho_k}{2 (1 - \rho_k)^2}\right) + 1 \\
        &<\frac{7}{2} \frac{\rho_k}{(1-\rho_k)} + \frac{2}{\Log(1/\rho_k)} \Log \left(\frac{1}{1 - \rho_k}\right) + 1.
    \end{align*}
Note that
\begin{align*}
    \frac{2}{\Log(1/\rho_k)} \Log \left(\frac{1}{1-\rho_k}\right) = 
    \frac{2}{\Log(1/\rho_k)} \left( \Log\left( \frac{\rho_k}{1 - \rho_k}\right) + \Log  \left(\frac{1}{\rho_k}\right)  \right) = \frac{2}{\Log(1/\rho_k)}  \Log\left( \frac{\rho_k}{1 - \rho_k}\right) + 2,
\end{align*}
hence
\begin{equation*}
    \expect{n_k(T)} < \frac{7}{2} \frac{\rho_k}{(1-\rho_k)} + \frac{2}{\Log(1/\rho_k)}  \Log\left( \frac{\rho_k}{1 - \rho_k}\right) + 3.
\end{equation*}
    We recall from \autoref{thm:dbounds} the inequalities
    \begin{equation*}
        \frac{\rho_k}{1- \rho_k} \leq \frac{1}{\Log\left(1/\rho_k\right)} < \left\{\begin{array}{cc}
            D, & \Delta_k > 1, \\
            D /\Delta_k^2, & \Delta_k \leq 1.
        \end{array} \right.
    \end{equation*}
    Thus, 
\begin{equation*}
    \expect{n_k(T)} < \left\{\begin{array}{cc}
            7D/2 + 2D \Log D + 3, & \Delta_k > 1, \\
            7D/(2\Delta_k^2) + 2D/\Delta_k^2 \Log \left( D/\Delta_k^2 \right) + 3, & \Delta_k \leq 1.
        \end{array} \right.
\end{equation*}
and therefore
\begin{equation*}
    \expect{n_k(T)} < 
            \frac{7 D}{2} \left(1 + \frac{1}{\Delta_k^2}\right) + 2D\left(\Log D + \frac{1}{\Delta_k^2} \Log \left( D/\Delta_k^2 \right)\right) + 9,
\end{equation*}
which proves the result given an additional multiplication by $\Delta_k$. This completes the proof.
\end{proof}

\subsection{Proof of the Main Result}
\label{sec:mainresultproof}
In this section we prove our estimation of the Bayesian regret given certain conditions on the prior. To do this we first need two lemmas that divides the \hyperref[thm:mainlemma]{Main Lemma} into two parts and bounds them separately. We then conclude this section with a proof of \autoref{thm:main}.
% \todo[color=orange]{Fixed.}
% \todo{Avnittet är en blandning mellan resonemang och lemman. Tänkbart upplägg:Först metatext om att i detta avsnitt bevisar vi teorem 2.1 och att beviset kommer efter följande två lemma. Efter dessa kan man sedan kommer lägga begin{proof}[Proof of Theorem 2.1] (eller liknande)}

Thus, we now leave the confines of the single environment view and start to regard both the immediate regret $\Delta_k = \mu^* - \mu_k$ and any derived quantities as random variables under the Bayesian prior. Recall from the \hyperref[thm:mainlemma]{Main Lemma} that for $T \geq 2$ we have the arm specific regret bound
\begin{equation*}
    \condexpect{n_k(T)}{\theta} \Delta_k < 
            \frac{7 D}{2} \left(\Delta_k + \frac{1}{\Delta_k}\right) + 2D\left(\Log D \Delta_k + \frac{1}{\Delta_k} \Log \left(\frac{D}{\Delta_k^2} \right)\right) + 9 \Delta_k.
\end{equation*}
Dividing the bound into a sum of a non-reciprocal part
\begin{equation}
\label{eq:hu}
    h_{u}(\Delta_k) \coloneqq \frac{7 D}{2} \Delta_k + 2 D \Log D \Delta_k + 9 \Delta_k,
\end{equation}
and a reciprocal part 
\begin{equation}
\label{eq:hl}
    h_{l}(\Delta_k) \coloneqq \left(\frac{7 D}{2} + 2 D \Log D \right) \frac{1}{\Delta_k} + 2 D \frac{1}{\Delta_k} \Log \left(\frac{1}{\Delta_k^2}\right),
\end{equation}
implies the following bound on the Bayesian regret in \eqref{eq:bayesregret}: 
\begin{equation}
\label{eq:huhl}
    \expect{\sum_{k \in [K]} n_k(t) \Delta_k} <  \expect{\sum_{k \in [K]} h_{u}(\Delta_k)} + \expect{\sum_{k \in [K]} h_{l}(\Delta_k)}.
\end{equation}
We also recall that the Bayesian prior dictates that model parameters $\left(\theta_k, \tau_k \right)_{k = 1}^K$ are independently drawn from $\normalgamma\left(0, \precision_k^*, \alpha^*, \beta^* \right)$, $\alpha^* > 5/2$, $k \in [K]$. So for any normalized context $a_k \in \mathbb{R}^d$ we have
\begin{align*}
\lambda_k^* &\coloneqq \left(a_k^\transpose \left(\precision_k^*\right)^{-1} a_k\right)^{-1}, \\
    \mu_k &\sim \mathcal{N}\left(0, (\lambda_k^* \tau_k)^{-1} \right), \\
    \tau_k &\sim \Gam\left(\alpha^*, \beta^*\right),
\end{align*}
such that $(\mu_k, \tau_k) \sim \normalgamma\left(0, \lambda_k^* , \alpha^*, \beta^*\right)$. Moreover, by \autoref{def:cd} we have $\tau_0 \coloneqq \min_k \tau_k$ and 
\begin{align*}
    C(\tau_0) &\coloneqq \frac{2}{\tau_0} \sqrt{\frac{ \tau_0}{2} + \Log 2} + \frac{1}{\tau_0} \left(2 \Log 2 + 1 \right) + 3, \\
    D(\tau_0) &\coloneqq \frac{8}{\tau_0}\left(1 + \sqrt{\frac{\tau_0 \left(2 C(\tau_0) + \frac{1}{4} \right)}{2}} \right)^2.
\end{align*}
It is not hard to see that $D(\tau_0) > 1$. In addition, if $\tau_0 > 1$ then $D = \mathcal{O}(1)$, and if $\tau_0 \leq 1$ then $D = \mathcal{O}(1/\tau_0)$. Hence, $D = \mathcal{O}(1 + 1/\tau_0)$ and $D^2 = \mathcal{O}\left(1 +  1/\tau_0^2\right)$. 

Thus by the premises of the Bayesian prior and by analyzing \eqref{eq:hu} we obtain the following result.
\begin{lemma}
\label{thm:hu}
Suppose that $\gamma$ satisfies $5/2 < \gamma < \alpha^*$. If we let $\epsilon \coloneqq 1/\gamma$ then
\begin{equation*}
    \expect{\sum_{k \in [K]} h_{u}(\Delta_k)} = \mathcal{O}\left( K^{1 + 5 \epsilon/2} \sqrt{\Log K} \right).
\end{equation*}
\end{lemma}
\begin{proof}
We have 
\begin{equation*}
     h_{u}(\Delta_k) \coloneqq \frac{7 D}{2} \Delta_k + 2 D \Log D \Delta_k + 9 \Delta_k.
\end{equation*}
In view of \autoref{thm:deltabounds}, \autoref{thm:boinvgamma} and the fact that $D = \mathcal{O}(1 + 1/\tau_0)$ and $D^2 = \mathcal{O}\left(1 +  1/\tau_0^2\right)$, we obtain
\begin{equation*}
\begin{aligned}
    \expect{\Delta_k} &= \mathcal{O}(\expect{1/\tau_0^{1/2}} \sqrt{\Log K}) = \mathcal{O}\left( K^{\epsilon/2} \sqrt{\Log K} \right), \\
    \expect{D \Delta_k} &= \mathcal{O}(\expect{1/\tau_0^{3/2}} \sqrt{\Log K}) = \mathcal{O}\left( K^{3\epsilon/2} \sqrt{\Log K} \right), \\
     \expect{D \Log D \Delta_k} &< \expect{D^2 \Delta_k} = \mathcal{O}(\expect{1/\tau_0^{5/2}} \sqrt{\Log K}) = \mathcal{O}\left( K^{5\epsilon/2} \sqrt{\Log K} \right).
\end{aligned}
\end{equation*}
It now follows that 
\begin{equation*}
    \expect{\sum_{k \in [K]} h_{u}(\Delta_k)} = \sum_{k \in [K]} \expect{\frac{7 D}{2} \Delta_k + 2 D \Log D \Delta_k + 9 \Delta_k} = \mathcal{O}\left( K^{1 + 5 \epsilon/2} \sqrt{\Log K} \right)
\end{equation*}
as required. This concludes the proof.
\end{proof}
We now turn our attention to the reciprocal expression in \eqref{eq:hl}. By bounding the combined total of all arms for a specific environment we may in turn achieve a bound on the Bayesian regret. Explicitly we have the following result.
\begin{lemma}
\label{thm:hl}
Suppose that $\gamma$ satisfies $5/2 < \gamma < \alpha^*$. If we let $\epsilon \coloneqq 1/\gamma$ then for $T \geq 2$,
\begin{equation*}
    \expect{\sum_{k \in [K]} h_{l}(\Delta_k)} = \mathcal{O}\left(K^{3 \epsilon}\sqrt{\frac{K T}{ W_0\left(\displaystyle \frac{T}{K^{1 - 2\epsilon}}\right)}}  +  \sqrt{K T W_0\left(\displaystyle \frac{T}{K^{1 - 2\epsilon}}\right)}\right),
\end{equation*}
where $W_0(x)$ is Lambert's principal $W$-function.
\end{lemma}
\begin{proof}
We recall from \eqref{eq:hl} that we have
\begin{equation}
\label{eq:invdelta}
        h_{l}(\Delta_k) \coloneqq \left(\frac{7 D}{2} + 2 D \Log D \right) \frac{1}{\Delta_k} + 2 D \frac{1}{\Delta_k} \Log \left(\frac{1}{\Delta_k^2}\right).
\end{equation}
Note that Lambert's principal $W$-function $W_0$ satisfies
\begin{equation*}
    x = W_0(x) \exp\left(W_0(x)\right) \quad (x > 0)
\end{equation*}
such that $\exp\left(-W_0(x)\right) = W_0(x)/x$. 

Let us fix an environment and put $\delta \coloneqq \exp\left(-W_0\left(\displaystyle \frac{T}{K^{1 - 2\epsilon}}\right)/2\right)$. This implies the identity
\begin{equation}
\label{eq:epsidentity}
    \frac{1}{\delta^2} \Log \left( \frac{1}{\delta^2}\right) = W_0\left(\frac{T}{K^{1 - 2\epsilon}}\right) \exp\left(W_0\left(\frac{T}{K^{1 - 2\epsilon}}\right) \right) = \frac{T}{K^{1 - 2\epsilon}}.
\end{equation}
It follows that if we consider arms where $\Delta_k \leq \delta$ we have $\delta T$ as the maximum possible combined regret contribution from all such arms. Moreover, since $\delta = \sqrt{K^{1 - 2\epsilon} W_0\left(\displaystyle \frac{T}{K^{1 - 2\epsilon}}\right) / T }$ we may express this as
\begin{equation}
\label{eq:goodarms}
    \delta T = \sqrt{\frac{K^{1 - 2\epsilon} W_0\left(\displaystyle \frac{T}{K^{1 - 2\epsilon}}\right)}{T}} T = \sqrt{K^{1 - 2\epsilon} T W_0\left(\displaystyle \frac{T}{K^{1 - 2\epsilon}}\right)}.
\end{equation}
On the other hand, if $\Delta_k > \delta$ then $\frac{1}{\Delta_k} < \frac{1}{\delta}$. Hence in view of \eqref{eq:invdelta} and \eqref{eq:epsidentity} we have the total contribution from these arms bounded by
\begin{align}
    &\sum_{k = 1}^K \left(\frac{7 D}{2} + 2 D \Log D \right) \frac{1}{\Delta_k} + 2 D \frac{1}{\Delta_k} \Log \left(\frac{1}{\Delta_k^2}\right) < \left(\frac{7 D}{2} + 2 D \Log D\right) \frac{K}{\delta} + 2 D K \delta \frac{1}{\delta^2} \Log\left(\frac{1}{\delta^2}\right) \nonumber \\
    &= \left(\frac{7 D}{2} + 2 D \Log D\right) K \sqrt{\frac{T}{K^{1 - 2\epsilon} W_0\left(\displaystyle \frac{T}{K^{1 - 2\epsilon}}\right)}} + 2 D \delta T \nonumber \\
    &=\left(\frac{7 D}{2} + 2 D \Log D\right) \sqrt{\frac{K^{1 + 2 \epsilon} T}{ W_0\left(\displaystyle \frac{T}{K^{1 - 2\epsilon}}\right)}}  + 2 D \sqrt{K^{1 - 2\epsilon} T W_0\left(\displaystyle \frac{T}{K^{1 - 2\epsilon}}\right)}. \label{eq:trickyarms}
\end{align}
So by combining \eqref{eq:goodarms} and \eqref{eq:trickyarms} for a fixed environment we obtain the bound 
\begin{equation}
\label{eq:hlbound}
    \sum_{k \in [K]} h_l(\Delta_k) < \left(\frac{7 D}{2} + 2 D \Log D\right) \sqrt{\frac{K^{1 + 2 \epsilon} T}{ W_0\left(\displaystyle \frac{T}{K^{1 - 2\epsilon}}\right)}}  + (2 D + 1) \sqrt{K^{1 - 2\epsilon} T W_0\left(\displaystyle \frac{T}{K^{1 - 2\epsilon}}\right)}
\end{equation}
in terms of the random variable $D$.

We recall the fact that $D = \mathcal{O}(1 + 1/\tau_0)$ and $D^2 = \mathcal{O}\left(1 +  1/\tau_0^2\right)$. So in view of \autoref{thm:boinvgamma} we obtain
\begin{equation*}
    \begin{aligned}
        \expect{D} &= \mathcal{O}(\expect{1/\tau_0}) = \mathcal{O}(K^\epsilon), \\
        \expect{D \Log D} &< \expect{D^2} = \mathcal{O}(\expect{1/\tau_0^2}) = \mathcal{O}(K^{2\epsilon}).
    \end{aligned}
\end{equation*}
Thus by taking the expectation of \eqref{eq:hlbound} we obtain 
\begin{align*}
    &\expect{ \sum_{k \in [K]} h_l(\Delta_k) } < \expect{\left(\frac{7 D}{2} + 2 D \Log D\right) \sqrt{\frac{K^{1 + 2 \epsilon} T}{ W_0\left(\displaystyle \frac{T}{K^{1 - 2\epsilon}}\right)}}  + (2 D + 1) \sqrt{K^{1 - 2\epsilon} T W_0\left(\displaystyle \frac{T}{K^{1 - 2\epsilon}}\right)}} \nonumber \\
    &= \mathcal{O}\left(K^{3 \epsilon}\sqrt{\frac{K T}{ W_0\left(\displaystyle \frac{T}{K^{1 - 2\epsilon}}\right)}}  +  \sqrt{K T W_0\left(\displaystyle \frac{T}{K^{1 - 2\epsilon}}\right)}\right)
\end{align*}
as required. This completes the proof.
\end{proof}

We now have everything we need to prove the main result in \autoref{thm:main}. 
\begin{proof}[Proof of \autoref{thm:main}]
With summation of the corresponding bounds of \autoref{thm:hu} and \autoref{thm:hl} for the constituent parts of \eqref{eq:huhl} we can conclude that the Bayesian regret is bounded as
\begin{equation}
\label{eq:finalregret}
    \mathcal{O}\left(K^{3 \epsilon}\sqrt{\frac{K T}{ W_0\left(\displaystyle \frac{T}{K^{1 - 2\epsilon}}\right)}}  +  \sqrt{K T W_0\left(\displaystyle \frac{T}{K^{1 - 2\epsilon}}\right)} + K^{1 + 5 \epsilon/2} \sqrt{\Log K} \right). 
\end{equation}
It is not hard to see that if $x > e$, then $W_0(x) > 1$ and $W_0(x) < \Log x$. So in particular for sufficiently large $T > e K^{1-2\epsilon}$ the regret bound in \eqref{eq:finalregret} is 
\begin{equation*}
    \mathcal{O}\left(\sqrt{K T W_0\left(\displaystyle \frac{T}{K^{1 - 2\epsilon}}\right)}\right),
\end{equation*}
where $W_0\left(\displaystyle \frac{T}{K^{1 - 2\epsilon}}\right) < \Log \left(\displaystyle \frac{T}{K^{1 - 2\epsilon}}\right)$.
\end{proof}

\section{Discussion}
\label{sec:discussion}
We analyzed the Thompson sampling algorithm for a class of stochastic linear bandits with finitely many independent arms, where the reward for each arm can be seen as a normal random variable that depends linearly on an unknown parameter vector in $\mathbb{R}^d$ and with unknown variance. We showed that with normal-gamma priors on environment uncertainty one may attain near-optimal Bayesian regret bounds that are similar to, but sharper than the well-known result of order $\sqrt{K T \Log T}$ for TS-algorithms over bounded rewards. It should be noted that the argument used to prove \autoref{thm:hl} only takes into account prior information about model precisions. Thus, future work readily extends to further analysis of conditional expectations involving the reciprocal immediate regret. 

% In fact with a fixed environment we conjecture that $\expect{\frac{1}{\Delta_k} \Log \left(\frac{1}{\Delta_k}\right) \mid \Delta_k > \delta} = \mathcal{O}(K \Log^2 1/\delta)$. Minimizing $f(K) \Log^2 1/\delta + T \delta$ would then imply a regret bound of order $f(K) W_0^2(T/(2 f(K))$, which is similar to the gracefully increasing Bayesian risk of order $K^2 \Log^2 T$ computed for the active exploration policy by \textcite[Theorem~4.2]{rusmevichientong2010linearly}.

\printbibliography

\appendix
\section{Additional Facts and Inequalities}
In the analysis we utilize high and low probability events to bound the expected regret of \autoref{alg:ngts}. One such event is to pose the question if the sample mean of a Gaussian variable lies sufficiently above a fixed limit around the mean. 

Let $\varphi$ and $\Phi$ denote the density and cdf of the standard normal distribution. Put 
$$\mathcal{Q}(x) \coloneqq 1 - \Phi(x) = \Phi(-x)$$
as the complementary cdf. Then for clarity we recall some well-known facts concerning Gaussian distributions.
\begin{lemma}
\label{thm:gaussbounds}
Let $X \sim \mathcal{N}(\mu, \sigma^2)$ and let $c \geq 0$. Then we have
\begin{enumerate}[(i)]
    \item $\prob{X > \mu - c\sigma} = \Phi(c)$,
    \item $\prob{X > \mu + c\sigma} = \mathcal{Q}(c)$,
\item $\mathcal{Q}(c) \leq \frac{1}{2} \exp\left(-\frac{c^2}{2}\right)$.
\end{enumerate}
In particular, $\prob{X > \mu - c\sigma}$ is lower bounded by $1 - \frac{1}{2} \exp\left(-\frac{c^2}{2}\right)$.
\end{lemma}
\begin{proof}
The first two statements follow immediately from the fact that the cdf of $X$ is given by $\Phi\left(\frac{x - \mu}{\sigma}\right)$. Hence 
\begin{align*}
    \prob{X > \mu - c\sigma} &= 1 - \Phi\left(-c\right) = \Phi(c), \\
    \prob{X > \mu + c\sigma} &= 1 - \Phi\left(c\right) = \mathcal{Q}(c).
\end{align*}
For the third statement we define the continuous function
\begin{equation*}
  f(c) \coloneqq \frac{1}{2} \exp\left(-\frac{c^2}{2}\right) - \mathcal{Q}(c) = \frac{1}{2} \exp\left(-\frac{c^2}{2}\right) + \Phi(c) - 1.
\end{equation*}
Then $f(0) = f(\infty) = 0$ at the boundaries. Moreover, the equation
\begin{align*}
    f'(c) &= -\frac{c}{2} \exp\left(-\frac{c^2}{2}\right) + \varphi(c) \\
    &=\exp\left(-\frac{c^2}{2}\right) \left(- \frac{c}{2} + \frac{1}{\sqrt{2 \pi}}\right) = 0
\end{align*}
yields a single critical point at $c_0 = \sqrt{\frac{2}{\pi}}$, where
\begin{equation*}
    f(c_0) = \frac{1}{2} \exp \left(-\frac{1}{\pi}\right) + \Phi\left(\sqrt{\frac{2}{\pi}}\right) - 1 > 0.
\end{equation*}
So $f(c_0) > 0$ combined with the boundary conditions implies $f(c) > 0$ for all $c \in (0, \infty)$. Hence $\mathcal{Q}(c) \leq \frac{1}{2} \exp\left(-\frac{c^2}{2}\right)$ for all non-negative $c$.

The lower bound now follows from
\begin{equation*}
    \prob{X > \mu - c\sigma} = \Phi(c) = 1 - \mathcal{Q}(c) \geq 1 - \frac{1}{2} \exp\left(-\frac{c^2}{2}\right).
\end{equation*}
This completes the proof.
\end{proof}
Moreover, in the analysis we will also need some fairly straight forward inequalities concerning the natural logarithm.
\begin{lemma}
\label{thm:logprops}
For any $x\geq 0$,
\begin{equation*}
    \Log(1+x) \geq \frac{2 x}{2 + x}.
\end{equation*}
In addition, if $0 < x < 1$ then 
\begin{equation*}
    \frac{1}{\Log\left(\frac{1}{x}\right)} \geq \frac{x}{1-x}.
\end{equation*}
\end{lemma}
\begin{proof}
We have
\begin{equation*}
    \Log(1 + x) = \int_{0}^x \frac{1}{1+t} \ d t,
\end{equation*}
where we note that the derivative $f(t) = \frac{1}{1 + t}$ is convex on $(0,x)$. Hence, $f(t)$ is bounded below by its corresponding linearization $L(t ; t_0)$ at $t_0 = x/2$. Explicitly,
\begin{equation*}
   L(t; t_0) = f(t_0) + f'(t_0)(t-t_0) = \frac{1}{1 + t_0} - \frac{t - t_0}{(1+t_0)^2} = \frac{1 + 2 t_0 - t}{(1 + t_0)^2}.
\end{equation*}
Thus
\begin{align*}
    \Log(1+x) \geq \int_0^x L(t;t_0) \ d t = \frac{2 x}{2 + x}
\end{align*}
as required.

For the last statement, put $w \coloneqq 1/\Log\left(1/x\right)$ such that $1/x = e^{\frac{1}{w}}$. Note that for any positive $y$ we have $e^y - 1 \geq y$. Thus
\begin{equation*}
    \frac{1}{x} - 1 = e^{\frac{1}{w}} - 1 \geq \frac{1}{w}
\end{equation*}
or equivalently
\begin{equation*}
    \frac{1}{\Log\left(\frac{1}{x}\right)} = w \geq \frac{x}{1 - x}.
\end{equation*}
This completes the proof.
\end{proof}

\end{document}